\documentclass[11pt]{article}

\usepackage[final]{acl}

\usepackage{times}
\usepackage{latexsym}

\usepackage[T1]{fontenc}

\usepackage[utf8]{inputenc}

\usepackage{microtype}

\usepackage{inconsolata}

\usepackage{graphicx}
\usepackage{float}
\usepackage{xspace}
\usepackage{supertabular}
\usepackage{algorithm}
\usepackage{algpseudocode}
\usepackage{booktabs}
\usepackage{url}
\usepackage{makecell}
\usepackage{pifont,tikz}
\usepackage{xcolor} 
\usepackage{hyperref}
\usepackage{amsmath, amssymb, amsfonts}
\usepackage{multirow}
\usepackage[most]{tcolorbox}
\usepackage{fontawesome5}

\newtcolorbox{promptbox}[1][]{%
  enhanced,
  colback=gray!5,
  colframe=gray!50,
  boxrule=0.4pt,
  arc=2pt,
  left=6pt, right=6pt, top=4pt, bottom=4pt,
  fontupper=\small,
  breakable,
  #1
}
\definecolor{iconMath}{RGB}{255,140,0}      
\definecolor{iconBrain}{RGB}{147,51,234}    
\definecolor{iconCompMath}{RGB}{220,38,38}  
\definecolor{iconFinance}{RGB}{16,163,127}  
\definecolor{iconPython}{RGB}{55,118,171}   
\newcommand{\benchicon}[2]{{\color{#1}\makebox[1.3em][c]{#2}}} 
\newcommand{\logoicon}[1]{\raisebox{-0.2\height}{\includegraphics[height=1.2em]{figs/icons/#1}}}

\newcommand{\tkrNVDA}{\logoicon{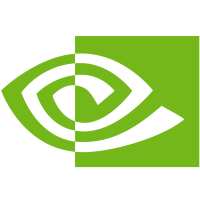}}
\newcommand{\tkrMSFT}{\logoicon{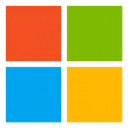}}
\newcommand{\tkrAAPL}{\logoicon{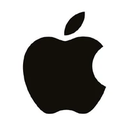}}
\newcommand{\tkrNFLX}{\logoicon{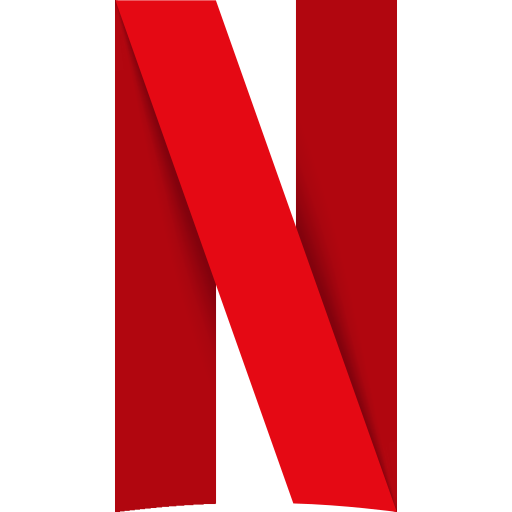}}
\newcommand{\tkrAMZN}{\logoicon{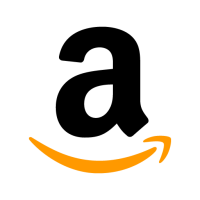}}

\newcommand{\llmPhi}{\logoicon{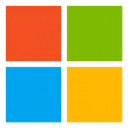}}
\newcommand{\llmQwen}{\logoicon{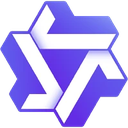}}
\newcommand{\llmLlama}{\logoicon{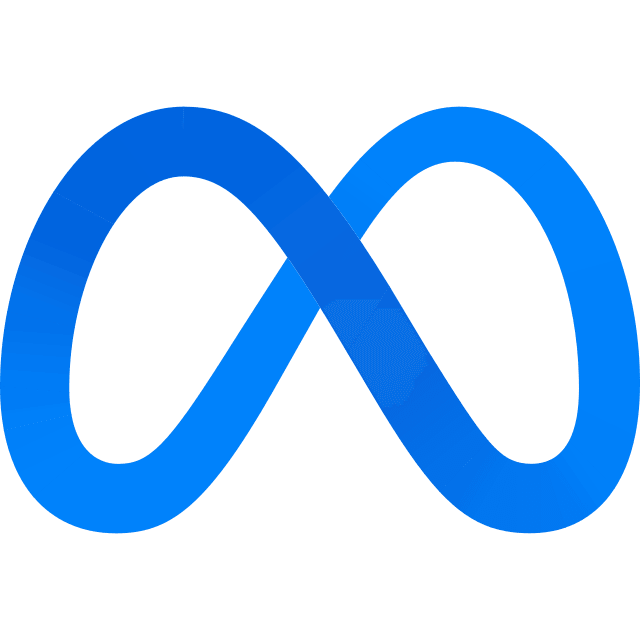}}
\newcommand{\llmStarling}{\logoicon{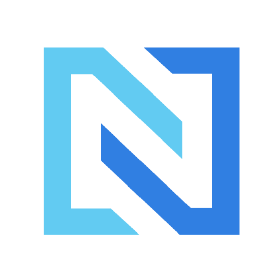}}
\newcommand{\llmDeepSeek}{\logoicon{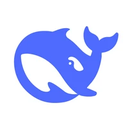}}
\newcommand{\llmMistral}{\logoicon{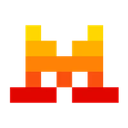}}
\newcommand{\llmYi}{\logoicon{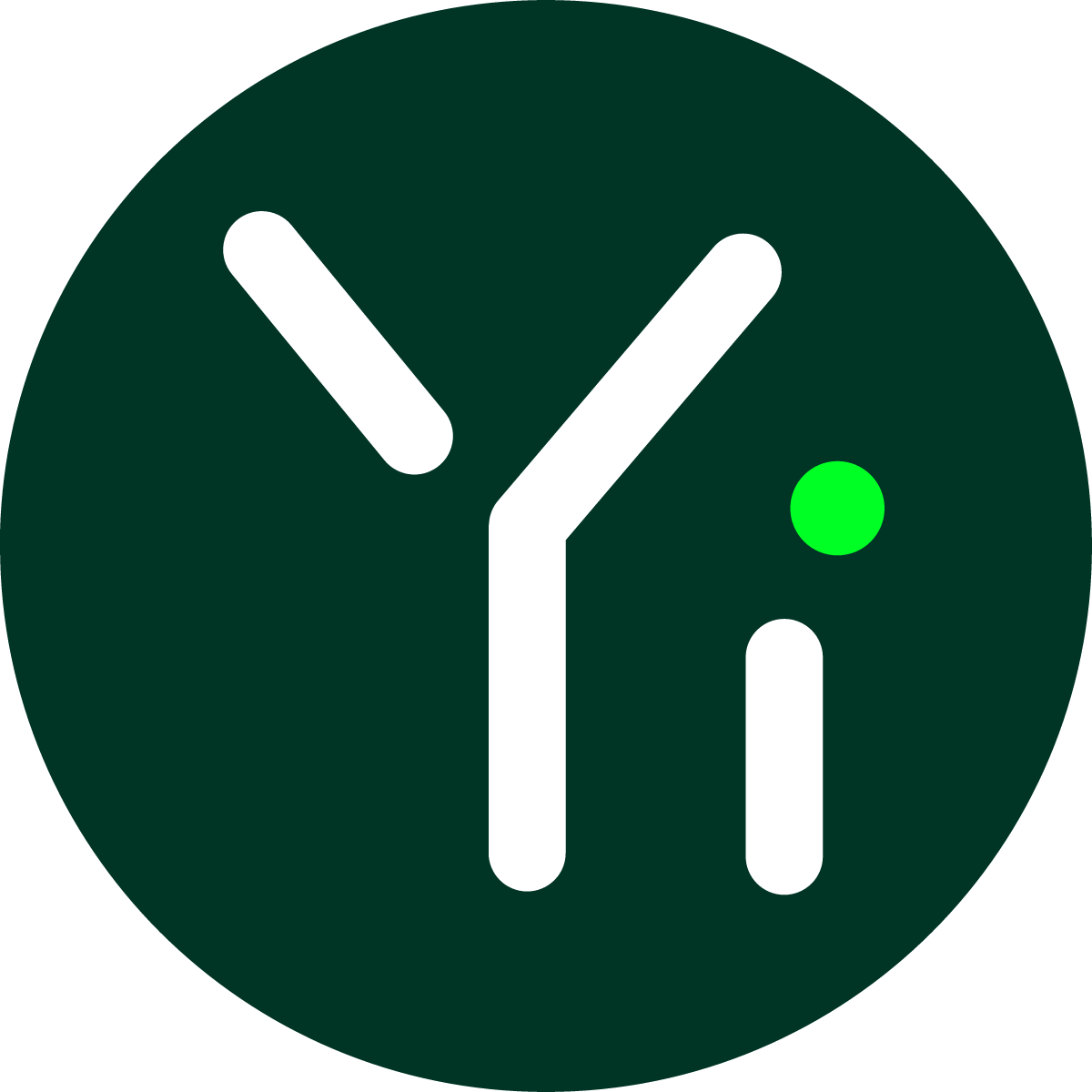}}
\newcommand{\llmOLMo}{\logoicon{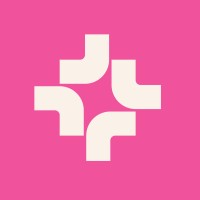}}
\newcommand{\llmFalcon}{\logoicon{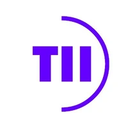}}
\newcommand{\llmGLM}{\logoicon{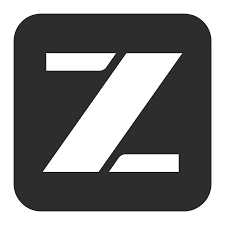}}

\newcommand{\ourmethod}{\textsc{FinCAD}\xspace}

\title{Summoning the Oracle to Slay It: Mitigating Look-Ahead Bias in Financial Backtesting with Large Language Models}

\author{
    Weixian Waylon Li\\
    University of Edinburgh \\
    {\texttt{waylon.li@ed.ac.uk}} \\\And
    Mengyu Wang \\
    University of Edinburgh \\
    {\texttt{mengyu.wang@ed.ac.uk}} \\ \And 
    Tiejun Ma \\
    University of Edinburgh \\
    {\texttt{tiejun.ma@ed.ac.uk}}
}

\begin{document}
\maketitle
\begin{abstract}
    Backtesting large language models (LLMs) on historical financial data is unreliable when their pre-training data include the evaluated events.
    An LLM trained in 2024 may already encode how stocks moved during 2018--2020.
    We name this failure \emph{parametric look-ahead bias} and propose \ourmethod, an inference-time adaptation of Context-Aware Decoding that attenuates contributions from memorised historical outcomes without retraining.
    \ourmethod pairs an adversarial bias-discovery pipeline that learns a model-specific memory-activating prior prompt with an entity- and date-adaptive rule that scales the CAD strength using a per-(entity, date) confidence signal.
    Across five 7--14B LLMs and five mega-cap equities, the largest model-level mean in-sample return correction is $-$67.1\%. For the three larger models, 2025 out-of-sample returns remain within \$8K and mean Sharpe within $\pm$0.10 of baseline; mean general-benchmark accuracy remains positive or within $-$1.7 points for four of five models.
    On an eleven-model leaderboard, \ourmethod raises the subset-averaged in-sample/out-of-sample Spearman correlation from $+0.779$ to $+0.846$, yielding rankings that are more closely aligned with post-cutoff performance.

\end{abstract}

\section{Introduction}
\label{sec:intro}
Large language models (LLMs) are increasingly applied as autonomous financial agents that read news, filings, and prices and decide whether to buy, sell, or hold \citep{10.1145/3768292.3770387}.
In quantitative finance, backtests gate capital allocation, so a result inflated by undetected bias translates directly into capital misallocation and live-trading losses; standardised corrections exist for backtest overfitting and multiple testing \citep{harvey2014evaluating,bailey2014pseudomath,bailey2014deflated,lopezdeprado2018advances}, survivorship truncation \citep{heckman1979sample,brown1992survivorship}, and point-in-time data leakage \citep{li2025llmbasedfinancialinvestingstrategies,kong2026evaluatingllmsfinancerequires}.
However, limited work has studied the bias introduced by LLMs.
A state-of-the-art LLM with a 2025 cutoff has already seen how NVIDIA, Microsoft, and Netflix moved during 2010--2020, so any backtest overlapping the training window inherits an additional bias that lives \emph{inside} the model's weights and is invisible to data-pipeline audits. 
We call this failure mode \emph{parametric look-ahead bias}, and its consequence is a trap: uncorrected long-horizon backtests are systematically misleading, whereas the only bias-free alternative, the short post-cutoff period, is too limited to support statistically reliable conclusions \citep{li2025llmbasedfinancialinvestingstrategies}.

A natural starting point is Context-Aware Decoding (CAD)~\citep{shi-etal-2024-trusting}, which sharpens the contribution of the input by subtracting an off-context prior at the logit level.
CAD was originally proposed to suppress \emph{unfaithful} priors (hallucinations); ours is the opposite regime, where the prior the model carries about ``NVDA in 2018'' is not a hallucination but an accurate memory of the future that backtesting must explicitly reject.
Two challenges follow: $x_{\text{prior}}$ must reliably activate that memory rather than answer the question, and $\alpha$ must vary across entities and dates since memorisation is non-uniform on both axes, so a fixed $\alpha$ would simultaneously under-correct memorised cases and over-penalise context-grounded reasoning on out-of-sample dates.

We propose \ourmethod\footnote{Code: \url{https://github.com/waylonli/FinCAD}}, an inference-time decoding scheme that mitigates parametric look-ahead bias without retraining by explicitly eliciting a memory-enriched prior and subtracting it from the context-conditioned logits with an adaptive strength.
Conceptually, the discovery stage ``summons the oracle'' by making the historical prior explicit; the title's ``slaying'' refers to attenuating that prior's contribution during decoding.
An \emph{adversarial bias-discovery} pipeline optimises a model-specific instruction $T_{\text{prior}}^*$ that activates the model's historical-outcome recall on a held-out probe set, yielding an explicit subtractable prior.
An \emph{entity- and date-adaptive} rule for $\alpha$ compares the model's per-(entity, date) directional entropy against an entity-level baseline, so the CAD strength rises with date-variance and per-date excess confidence and attenuates when this signal is absent.

Our contributions are: 

\begin{itemize}
    \item We propose \ourmethod, the first decoding-time adaptation of CAD designed for mitigating parametric look-ahead bias, with an adversarial discovery pipeline for $x_{\text{prior}}$ and an entity- and date-adaptive rule for $\alpha$ (\S\ref{sec:discovery}, \S\ref{sec:adaptive-alpha}).
    \item Across five LLMs (7B--14B), five mega-cap equities, and five reasoning benchmarks, \ourmethod produces model-dependent in-sample (2010--2020) performance corrections while mean benchmark accuracy remains positive or within $- 1.7$ points for four of five models (\S\ref{sec:exp-1}, \S\ref{sec:exp-2}).
    \item \ourmethod improves the descriptive alignment between in-sample and post-cutoff model rankings: across eleven LLMs on the S\&P~500, the subset-averaged Spearman correlation between in-sample (2010--2020) and 2025 Sharpe rankings rises from $+0.779$ to $+0.846$ (\S\ref{sec:exp-3}).
\end{itemize}

\section{Related Work}\label{sec:related-work}

A growing line of work casts LLMs as autonomous financial investors \citep{guo2024largelanguagemodelbased,10.1145/3768292.3770387}, including a recent wave of dedicated trading agents \citep{yu2023finmemperformanceenhancedllmtrading,finagentzhang2024multimodalfoundationagentfinancial,yang2024finrobotopensourceaiagent,ding2024tradexpertrevolutionizingtradingmixture,yu2024fincon,10.1145/3627673.3679653,wang-etal-2024-modeling,wang-etal-2025-one,xiao2025tradingagentsmultiagentsllmfinancial,10.1145/3768623,fatouros2025marketsenseai20enhancingstock,li-etal-2025-fingear,li2026timelabelcontinuousphase} and reinforcement-learning-enhanced variants \citep{RePEc:arx:papers:2310.05627,10.1145/3589334.3645611,xiao2025tradingagentsmultiagentsllmfinancial,deng2026alphaquanterendtoendtoolaugmentedagentic}, most of which evaluate on short historical windows without bias control, leaving open whether reported gains reflect genuine reasoning or contamination of the evaluation regime.

Input-side protections (survivorship, data-snooping, data-leakage look-ahead) are tracked by FINSABER \citep{li2025llmbasedfinancialinvestingstrategies}, but none addresses the bias that resides in the model's weights themselves. 
This \emph{parametric look-ahead} is documented by \citet{benhenda2026lookaheadbenchstandardizedbenchmarklookahead}, who shows that LLMs recall in-cutoff S\&P~500 prices verbatim, and by \citet{gao2025testlookaheadbiasllm}, who provide a complementary statistical test; concurrently, \citet{merchant2026a} propose a decoding-time mitigation that fine-tunes a pair of auxiliary models on labelled forget/retain partitions, whereas \ourmethod uses a single base model with no auxiliary fine-tuning.

Our work connects to the broader literature on LLM memorisation \citep{carlini2021extracting,carlini2023quantifying,li2026spectral} and \emph{machine unlearning} for LLMs \citep{liu2024rethinking}, including parameter-editing methods \citep{meng2022locating,meng2023memit} and fine-tuning-based forget protocols \citep{jang2023knowledgeunlearning,eldan2023whosharrypotter,maini2024tofu}. 
All require curated forget sets, retraining cycles, or structured (subject, relation, object) edits, and risk collateral damage to neighbouring knowledge; they are operationally impractical at the eleven-model leaderboard scale of financial backtests, where each model would need a bespoke unlearning pass per ticker per evaluation date.
Decoding-side interventions instead operate at inference without modifying weights \citep{li-etal-2023-contrastive,chuang2024dola}, and \ourmethod belongs to this family.
We build specifically on Context-Aware Decoding \citep{shi-etal-2024-trusting}, whose contrast between with-context and without-context branches matches our setting: a trading agent's memory of how a stock actually moved should not leak into its reading of the financial context provided at decision time.

\section{Methodology}\label{sec:methodology}

\begin{figure*}[htbp]
    \centering
    \includegraphics[width=1.0\linewidth]{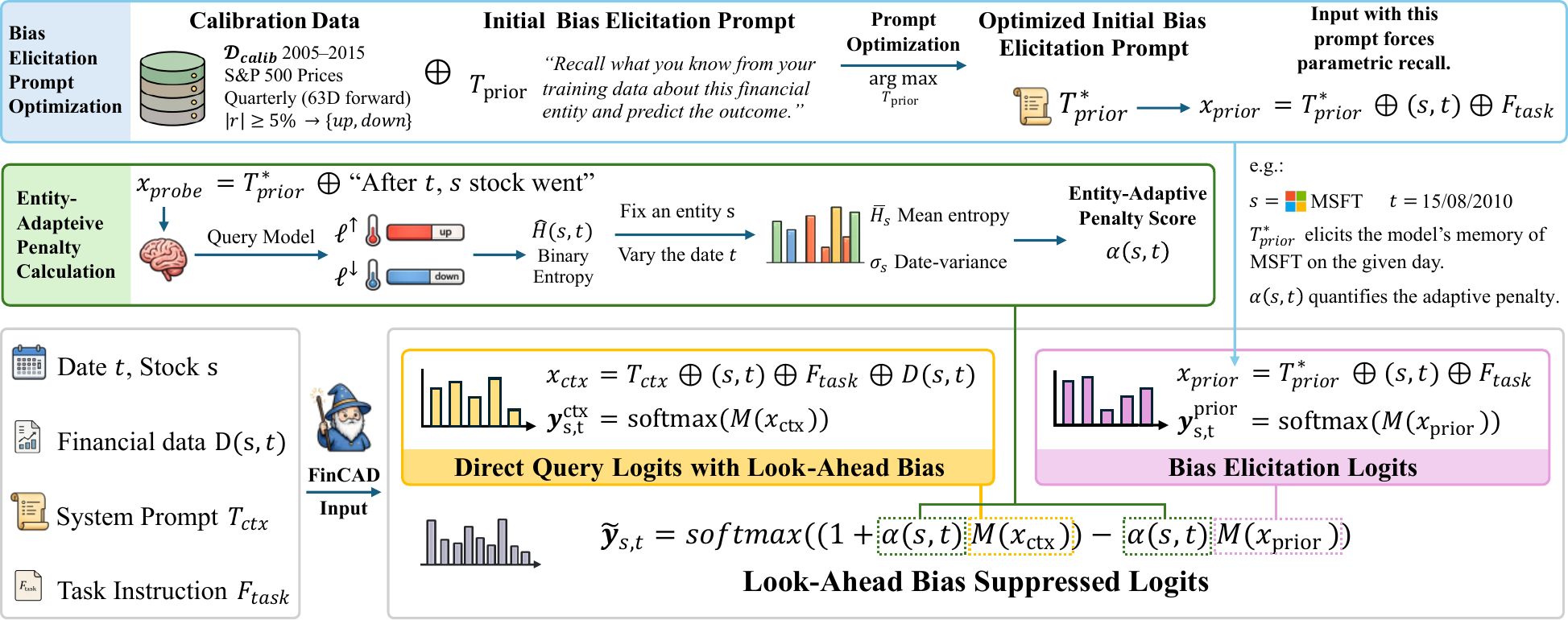}
    \caption{\ourmethod pipeline: \textbf{(1)} MIPROv2 discovers a memory-activation instruction $T_{\text{prior}}^*$ on $\mathcal{D}_{\text{calib}}$; \textbf{(2)} a completion probe and entity calibration set a per-(entity, date) strength $\alpha(s,t)$; \textbf{(3)} CAD subtracts the memory-enriched prior logits with strength $\alpha(s,t)$ from the context-conditioned logits.}
    \label{fig:method-fig}
\end{figure*}

Let $\mathcal{M}$ denote a language model with vocabulary $\mathcal{V}$.
For a given input string $x$, $\mathcal{M}(x) \in \mathbb{R}^{|\mathcal{V}|}$ denotes the model's next-token logit vector, and we write $\tilde{\boldsymbol{y}} = \mathrm{softmax}(\mathcal{M}(x))$ for the corresponding next-token output distribution over $\mathcal{V}$.
Context-Aware Decoding \citep{shi-etal-2024-trusting} replaces this ordinary output distribution with a contrastively adjusted distribution. 
In our \ourmethod, the adjustment strength varies per (entity, date) pair $(s,t)$, where $s$ is the stock symbol and $t$ the decision date:
\begin{multline}\label{eq:cad}
    \tilde{\boldsymbol{y}}_{s,t}
    = \mathrm{softmax}\!\bigl((1 + \alpha(s, t))\, \mathcal{M}(x_{\text{ctx}}) \\
    {} - \alpha(s, t)\, \mathcal{M}(x_{\text{prior}})\bigr)
\end{multline}
where $\tilde{\boldsymbol{y}}_{s,t}$ denotes the \ourmethod-adjusted next-token output distribution for entity $s$ at date $t$.
The original CAD targets unfaithful hallucinations and treats the prior as noise to be discounted. 
We repurpose it for the opposite regime, where a memory-enriched prior about historical outcomes is actively elicited and subtracted.
This leads to two design choices.
$x_{\text{prior}}$ must maximally activate that memory rather than answer the question; and $\alpha(s, t)$ should vary across both entities and time.
We address these in \S\ref{sec:discovery} and \S\ref{sec:adaptive-alpha} respectively; Figure~\ref{fig:method-fig} summarises the complete procedure, with full pseudocode in Algorithm~\ref{alg:fincad} (Appendix~\ref{app:algorithm}).

\subsection{Adversarial Bias Discovery}\label{sec:discovery}

The prior prompt $x_{\text{prior}}$ determines what knowledge is subtracted during decoding; if it fails to activate the model's memorised future knowledge, the subtraction has no effect and look-ahead bias persists, so we need a prior that reliably triggers parametric recall of historical outcomes.

For a decision over entity $s$ at date $t$, the two branches of Equation~\ref{eq:cad} share structural fields but differ in instruction and content. The context branch carries the full agent input:
\begin{equation}\label{eq:ctx-structure}
    x_{\text{ctx}} = T_{\text{ctx}} \;\oplus\; (s, t) \;\oplus\; F_{\text{task}} \;\oplus\; D(s, t)
\end{equation}
where $\oplus$ denotes string concatenation, $T_{\text{ctx}}$ is the system prompt framing the model as a quantitative analyst, $F_{\text{task}}$ is the task-specific instruction (e.g., ``Analyse the following financial data and decide whether to buy, sell, or hold'') together with the output-format specification (``Respond with valid JSON''), and $D(s, t)$ is the financial context the agent reasons over.
To realise the shared substructure with $x_{\text{prior}}$ at decoding time, $F_{\text{task}}$ is extracted and copied verbatim into the prior input below.
The prior branch strips $D(s, t)$ and replaces $T_{\text{ctx}}$ with an adversarially optimised memory-activation instruction $T_{\text{prior}}^*$ designed to activate parametric recall:
\begin{equation}\label{eq:prior-structure}
    x_{\text{prior}} = T_{\text{prior}}^* \;\oplus\; (s, t) \;\oplus\; F_{\text{task}}
\end{equation}
$T_{\text{prior}}^*$ is optimised once per model.
Sharing the $(s, t)$ and $F_{\text{task}}$ fields across the two branches keeps both inputs over the same structural tokens, so the logit subtraction in Equation~\ref{eq:cad} targets the \emph{content} of the response rather than its format or task framing. During autoregressive decoding, each sampled token is appended to both branches; every subsequent step therefore contrasts the same candidate tokens conditioned on a shared generated prefix.

\paragraph{Optimisation objective.}
We frame the search for $T_{\text{prior}}^*$ as a discrete prompt optimisation problem over a calibration dataset $\mathcal{D}_{\text{calib}}$ of historical (entity, date, outcome) triples.
Each example presents the model with an entity $s_i$ and date $t_i$ and asks for the direction label $d_i \in \{\texttt{up}, \texttt{down}\}$ from memory alone, with no financial context. The objective is:
\begin{equation}\label{eq:objective}
    \hat{d}_i = \operatorname*{arg\,max}_{d}\; \mathcal{M}\!\left(T_{\text{prior}} \,\oplus\, (s_i, t_i) \,\oplus\, F_{\text{task}}^{\text{calib}}\right)_d
\end{equation}
\begin{equation}
\label{eq:objective-T}
    T_{\text{prior}}^* = \operatorname*{arg\,max}_{T_{\text{prior}}} \sum_{i} \delta(\hat{d}_i,\, d_i) 
\end{equation}
where $d$ is a direction label drawn from $\{\texttt{up}, \texttt{down}\}$; for each label $d$, $\mathcal{V}_d \subset \mathcal{V}$ collects its tokenizer variants (casing and leading-space differences), and we write $\mathcal{M}(X)_d := \max_{v \in \mathcal{V}_d} \mathcal{M}(X)_v$ as shorthand for the maximum logit over those variants. The sum runs over all triples $(s_i, t_i, d_i) \in \mathcal{D}_{\text{calib}}$, $T_{\text{prior}}$ is the candidate memory-activation instruction being optimised, $F_{\text{task}}^{\text{calib}}$ is a fixed calibration task-shell that specifies the permitted output labels but contains \emph{no} prediction instruction, and $\delta(\hat{d}_i, d_i)$ equals $1$ when the predicted direction matches the ground-truth label and $0$ otherwise.
Crucially, only $T_{\text{prior}}$ is optimised; $F_{\text{task}}^{\text{calib}}$ remains fixed, and the optimised $T_{\text{prior}}^*$ is later reused at inference with a different (backtest) $F_{\text{task}}$ via Eq.~\ref{eq:prior-structure}.

\paragraph{Calibration dataset.}
We construct $\mathcal{D}_{\text{calib}}$ from historical S\&P~500 price data covering the period 2005--2015, chosen to fall well within the pre-training data of modern LLMs. For each ticker $s$, we sample dates at quarter-end frequency and compute the forward return over the subsequent 63 trading days (approximately one calendar quarter):$r_{s,t} = (p_{s,\, t+63} - p_{s,t})/(p_{s,t})$,
where $p_{s,t}$ is the split- and dividend-adjusted closing price. We discard examples where $|r_{s,t}| < 0.05$, since small returns yield ambiguous direction labels. The remaining examples are labelled $d_{s,t} = \texttt{up}$ if $r_{s,t} > 0$ and $\texttt{down}$ otherwise. We balance the two classes by downsampling the majority and cap the dataset at 200 examples, split 80/20 into training and validation.

\paragraph{Optimiser.}
We solve the optimisation in Equation~\ref{eq:objective-T} using MIPROv2 \citep{opsahl-ong-etal-2024-optimizing}, a multi-stage instruction optimisation algorithm from the DSPy framework \citep{khattab2024dspy}. MIPROv2 generates a pool of candidate instructions by prompting a proposer model, evaluates each candidate's accuracy on the training set, and selects the best-performing candidate on the held-out validation set. We initialise the search with the seed instruction: \emph{``Recall what you know from your training data about this financial entity and predict the outcome.''}

To keep $T_{\text{prior}}^*$ task-agnostic, we hide task-specific information from MIPROv2's proposer: training data (ticker names, dates, labels) and bootstrap traces are withheld, so the proposer sees only a generic program skeleton and the current seed.
This prevents the proposer from leaking task-specific output formats (e.g., ``predict whether the stock went up or down'') into $T_{\text{prior}}^*$.
The evaluation metric, however, uses the full calibration input (including the output labels in $F_{\text{task}}^{\text{calib}}$), so candidates are still scored on their ability to surface memorised knowledge.
The optimised $T_{\text{prior}}^*$ is serialised as a JSON artefact and loaded at inference time; the per-model $T_{\text{prior}}^*$ values appear in Appendix~\ref{app:prompt-tstar}.

\subsection{Entity-Adaptive Penalty}\label{sec:adaptive-alpha}

A fixed penalty $\alpha$ applies the same subtraction strength to every entity and time period. This is problematic for two reasons. First, parametric look-ahead bias is non-uniform across entities: the model has strong parametric recall for well-known stocks (e.g., AAPL, NVDA) but near-random predictions for obscure or recently listed companies.
Applying a high $\alpha$ to the latter subtracts noise from the context-conditioned logits, degrading output quality. Second, memorisation is non-uniform across time: the model has detailed knowledge of historical periods covered by its training data but no memorised prices for dates beyond its knowledge cutoff.
A date-agnostic probe would assign the same high $\alpha$ to a well-known entity regardless of whether the backtest date falls inside or outside the training window, penalising the prior even when it carries no look-ahead information.
We therefore set $\alpha$ dynamically per entity \emph{and} date, based on how confident the model is in its own memory.

\paragraph{Completion-based probing.}
Given entity $s$ and date $t$, we construct a completion probe by prepending the optimised instruction $T_{\text{prior}}^*$ to a natural-language prefix that elicits a directional continuation: $x_{\text{probe}} = T_{\text{prior}}^* \;\oplus\; \text{``After } t \text{, } s \text{ stock went''}$.

We run a single forward pass and read the per-label logits $\mathcal{M}(x_{\text{probe}})_d$ (the max-over-variants shorthand introduced in \S\ref{sec:discovery}):
\begin{equation}\label{eq:probe-logits}
    \ell_{\uparrow} = \mathcal{M}(x_{\text{probe}})_{\texttt{up}}, \quad \ell_{\downarrow} = \mathcal{M}(x_{\text{probe}})_{\texttt{down}}
\end{equation}
These directional logits determine only the scalar gate $\alpha(s,t)$; they are not mapped to the agent's \texttt{buy}/\texttt{sell}/\texttt{hold} actions. Action generation remains the same-token, full-vocabulary contrast in Equation~\ref{eq:cad}.
\paragraph{Date-variance scaling.}
We convert the extracted logits into a normalised binary entropy $\hat{H} \in [0, 1]$:
\begin{align}
    p_{\uparrow} &= \frac{\exp(\ell_{\uparrow})}{\exp(\ell_{\uparrow}) + \exp(\ell_{\downarrow})}, \quad p_{\downarrow} = 1 - p_{\uparrow} \\
    \hat{H} &= -\frac{1}{\log 2}\bigl(p_{\uparrow} \log p_{\uparrow} + p_{\downarrow} \log p_{\downarrow}\bigr) \label{eq:entropy}
\end{align}
where $\hat{H} = 0$ indicates full confidence and $\hat{H} = 1$ indicates maximum uncertainty.

A na\"ive approach would threshold on raw entropy: penalise when $\hat{H}$ is low and abstain when it is high.
However, raw entropy conflates two distinct sources of confidence.
The model may be confident because it has \emph{memorised the specific outcome} for entity~$s$ at date~$t$ (temporal memorisation, the target of bias removal), or because it holds a \emph{persistent belief} about $s$ regardless of $t$ (a brand prior, e.g., ``NVIDIA is a strong company'').
Subtracting a brand prior removes legitimate reasoning without addressing look-ahead bias.

We disentangle these two signals through a \textbf{date-variance analysis}.
Before each backtest, we probe entity~$s$ across a set of $N$ calibration dates $\{t_1, \ldots, t_N\}$ drawn from the 2006--2015 period and compute:
\begin{align}
    \bar{H}_s &= \frac{1}{N} \sum_{i=1}^{N} \hat{H}(s, t_i) \label{eq:entity-mean} \\
    \sigma_s  &= \sqrt{\frac{1}{N} \sum_{i=1}^{N} \bigl(\hat{H}(s, t_i) - \bar{H}_s\bigr)^2} \label{eq:entity-std}
\end{align}
The \textbf{date-variance} $\sigma_s$ measures how much the model's directional confidence fluctuates across dates for a \emph{fixed} entity.
When $\sigma_s$ is high, the model's confidence varies with the date, a pattern consistent with date-specific knowledge.
When $\sigma_s$ is low, the model is similarly confident across dates, a pattern consistent with a stable entity prior.

The entity-adaptive penalty combines two multiplicative factors:
\begin{align}
    s_\sigma &= \min\!\bigl(1,\; \sigma_s / \sigma_{\text{ref}}\bigr) \label{eq:date-var-scale} \\
    s_H &= \max\!\bigl(0,\; (\bar{H}_s - \hat{H}(s, t)) / \Delta_{\text{range}}\bigr) \label{eq:per-date-conf} \\
    \alpha(s, t) &= \max\!\Bigl(\alpha_{\min},\; \min\!\bigl(\alpha_{\max}\, s_\sigma\, s_H,\; \alpha_{\text{cap}}\bigr)\Bigr) \label{eq:adaptive-alpha}
\end{align}
\noindent where $\alpha_{\max}$ scales the nominal peak penalty, $\alpha_{\text{cap}}$ is a hard upper bound on the resulting strength, $\alpha_{\min}$ is a non-negative floor (set to $0$ in our experiments), and $\sigma_{\text{ref}}$ and $\Delta_{\text{range}}$ are model-specific hyperparameters from profiling (described in the ``Model-specific normalisation'' paragraph below).
The first factor (date-variance scale, Eq.~\ref{eq:date-var-scale}) attenuates $\alpha$ for entities whose confidence is stable across dates, precisely those driven by brand priors rather than memorisation.
The second factor (per-date confidence, Eq.~\ref{eq:per-date-conf}) activates the penalty only at dates where the model is \emph{more confident than its own average} for that entity.

This design has three key properties.
First, it is \textbf{entity-adaptive}: each entity's $\alpha$ is calibrated to its own memorisation profile, avoiding both over-subtraction on brand-prior-driven entities and under-subtraction on strongly memorised entities.
Second, it is \textbf{temporally selective}: even for a memorised entity, $\alpha > 0$ only at dates where the model is unusually confident, avoiding blanket subtraction across the entire backtest period.
Third, it is \textbf{temporally attenuating}: when post-cutoff dates produce entropy above $\bar{H}_s$, $\bar{H}_s - \hat{H} < 0$ and the penalty falls to $\alpha = 0$.

\paragraph{Entity calibration.}
At the start of each backtest, we probe the target entity across $N = 12$ quarterly dates from 2006--2015 and compute $\bar{H}_s$ and $\sigma_s$ (Eqs.~\ref{eq:entity-mean}--\ref{eq:entity-std}).
The calibration dates are fixed and entity-agnostic, and lie within the period used to optimise $T_{\text{prior}}^*$, so the probe measures the same $T_{\text{prior}}^*$-activated recall that decoding will subtract.

\paragraph{Model-specific normalisation.}
$\sigma_{\text{ref}}$ and $\Delta_{\text{range}}$ come from a one-time profiling pass that records the entropy distribution over representative (entity, date) pairs.
We set $\sigma_{\text{ref}}$ to the standard deviation of the post-cutoff entropy distribution and $\Delta_{\text{range}}$ to the in-sample entropy standard deviation.
Neither value uses target labels or backtest outcomes, but using post-cutoff model confidence makes this normalisation label-free and transductive. An inductive variant can instead estimate $\sigma_{\text{ref}}$ on an earlier post-cutoff calibration window and freeze it before evaluation on a later window; we do not evaluate that variant here.

For reporting, we summarise the per-(entity, date) penalty over a backtest window by its mean: $\bar{\alpha}_{\text{IS}}$ averages $\alpha(s,t)$ over all in-sample dates of a backtest, and $\bar{\alpha}_{\text{OOS}}$ is the analogous mean over out-of-sample dates.

\paragraph{Computational structure.}
\ourmethod splits the work into offline and online stages, which clarifies the cost profile. The offline work is done once per model ($T_{\text{prior}}^*$ optimisation, profiling of $\sigma_{\text{ref}}$ and $\Delta_{\text{range}}$) and once per entity before each backtest ($\bar{H}_s$ and $\sigma_s$ from twelve dated probes). The online work consists of one completion probe per (entity, date) decision to set $\alpha(s,t)$, followed at each decoding step by the two-branch logit blend in Eq.~\ref{eq:cad}. This adds one prior-branch forward pass per generated token plus one probe per (entity, date) decision.


\section{Experiments and Results}\label{sec:experiments}\label{sec:results}

Our evaluation is structured around a central concern: does \ourmethod mitigate look-ahead bias without broadly degrading the model's task performance?

A non-selective penalty could reduce backtest returns while also degrading tasks that require general reasoning.
The three experiments examine complementary properties. \textbf{Experiment~1} (\S\ref{sec:exp-1}, \emph{reasoning preservation}) measures changes on standard reasoning benchmarks under a fixed model-specific strength $\alpha_{\text{bench}}$. \textbf{Experiment~2} (\S\ref{sec:exp-2}, \emph{in-sample performance correction}) evaluates the entity- and date-adaptive penalty $\alpha(s,t)$ on in-sample and post-cutoff dates. \textbf{Experiment~3} (\S\ref{sec:exp-3}, \emph{ranking alignment}) measures whether the \ourmethod-adjusted in-sample model ranking is more closely aligned with the post-cutoff ranking.

Experiments~2 and~3 share a single-stock backtesting framework\footnote{\url{https://github.com/virattt/ai-hedge-fund}}: the LLM acts as a quantitative analyst, receives a structured price-derived summary at each trading date, and emits a \texttt{buy}/\texttt{sell}/\texttt{hold} signal that is executed at the next opening price under a long-only portfolio.
We extend the framework with notional transaction costs and a liquidity cap for realism; full details are deferred to Appendix~\ref{app:backtesting-framework}.
At inference, \ourmethod adds one prior-branch forward pass per generated token and one completion probe per (entity, date) decision; the remaining calibration is performed offline (\S\ref{sec:adaptive-alpha}).

\subsection{Experiment 1: Reasoning Preservation}\label{sec:exp-1}

We evaluate \ourmethod across five instruction-tuned LLMs spanning 7B--14B parameters: \textbf{Phi-4-14B}, \textbf{Qwen2.5-14B}, \textbf{Llama-3.1-8B}, \textbf{Starling-7B}, and \textbf{DeepSeek-7B-Chat}.
Each model is run on five standard benchmarks under both baseline decoding and \ourmethod: FinEval \citep{ke-etal-2025-demystifying} (financial multiple-choice across 20 subtasks), MMLU-Pro \citep{wang2024mmlupro}, GSM8K \citep{cobbe2021trainingverifierssolvemath}, Competition-Math \citep{hendrycks2021measuring}, and HumanEval \citep{chen2021evaluatinglargelanguagemodels}.
Because these benchmarks contain no dated financial entity on which to evaluate $\alpha(s,t)$, we apply the fixed model-specific strengths $\alpha_{\text{bench}}$ listed in Table~\ref{tab:general-benchmarks}.
We distinguish this benchmark setting from $\bar{\alpha}_{\text{IS}}$, which is the empirical mean of the adaptive strengths over dated backtest decisions.

Table~\ref{tab:general-benchmarks} reports the full five-model sweep.
Mean accuracy remains positive or within $- 1.7$ points of baseline for four of five models; DeepSeek-7B-Chat is the exception ($-4.3$ points), driven primarily by a $-21.3$-point change on GSM8K.
Table~\ref{tab:exp1-five-model-detail} of Appendix~\ref{app:per-model-detail} pairs these per-benchmark changes with the mean baseline and \ourmethod accuracy for all five models.

\begin{table}[t]
\centering
\footnotesize
\setlength{\tabcolsep}{2.0pt}
\begin{tabular}{lrrrrr}
\toprule
& \shortstack{\llmPhi\\[-2pt]\textbf{Phi}}
& \shortstack{\llmQwen\\[-2pt]\textbf{Qwen}}
& \shortstack{\llmLlama\\[-2pt]\textbf{Llama}}
& \shortstack{\llmStarling\\[-2pt]\textbf{Starl.}}
& \shortstack{\llmDeepSeek\\[-2pt]\textbf{DSeek}} \\
\midrule
$\alpha_{\text{bench}}$ & .18 & .81 & .26 & .62 & .90 \\
\midrule
\benchicon{iconMath}{\faCalculator}\,GSM8K  & +0.5 & +0.4 & $-$4.9 & $-$6.9 & $-$21.3 \\
\benchicon{iconBrain}{\faBookOpen}\,MMLU-Pro  & $-$0.2 & +2.1 & +6.6 & +0.2 & +2.4 \\
\benchicon{iconCompMath}{\faSuperscript}\,Comp. Math  & +0.8 & $-$2.0 & $-$7.8 & $-$1.0 & $-$4.6 \\
\benchicon{iconFinance}{\faChartLine}\,FinEval  & +7.1 & $-$0.6 & +0.2 & $-$1.0 & +2.8 \\
\benchicon{iconPython}{\faPython}\,HumanEval  & +3.0 & +0.6 & +0.6 & 0.0 & $-$0.6 \\
\midrule
\textit{mean $\Delta$} & \textbf{+2.2} & \textbf{+0.1} & \textbf{$-$1.0} & \textbf{$-$1.7} & \textbf{$-$4.3} \\
\bottomrule
\end{tabular}
\caption{General-benchmark accuracy changes (\ourmethod $-$ baseline, percentage points) at the fixed model-specific strength $\alpha_{\text{bench}}$. Columns denote Phi-4-14B, Qwen2.5-14B, Llama-3.1-8B, Starling-7B, and DeepSeek-7B-Chat. Greedy decoding, seed 42.}
\label{tab:general-benchmarks}
\end{table}

\subsection{Experiment 2: In-Sample Performance Correction}\label{sec:exp-2}

We first run the single-stock backtesting framework on five heavily memorised mega-cap equities (NVDA, MSFT, AAPL, NFLX, AMZN) for the same five models introduced in Experiment~1.
These tickers are widely represented in financial news, filings, and analyst coverage, increasing the likelihood that their histories appear in pre-training corpora. Their prominence also makes them common testbeds in recent LLM-trading work \citep{yu2023finmemperformanceenhancedllmtrading,ding2024tradexpertrevolutionizingtradingmixture,xiao2025tradingagentsmultiagentsllmfinancial}.
Each ticker is evaluated over both the in-sample period (2010--2020, where evaluated dates precede every model's training cutoff) and the post-cutoff period (calendar year 2025); Table~\ref{tab:honesty-drop} reports Phi-4-14B and Qwen2.5-14B, with the other three models in Appendix~\ref{app:per-model-detail}.
This paired design measures whether \ourmethod produces larger corrections on dates that may overlap pre-training data than on dates beyond each model's reported cutoff.

\paragraph{In-sample performance correction.}
For Qwen2.5-14B, \ourmethod reduces returns on \emph{all five} tickers with a mean relative drop of $-$45.4\% of baseline return and $-$25.6\% of baseline Sharpe, reaching $-$78.2\% (NVDA) and $-$60.8\% (MSFT) on the most heavily covered names.
The mean calibrated $\bar{\alpha}_{\text{IS}}=1.06$ confirms that the entity-adaptive penalty is firing strongly on the dates Qwen2.5-14B is most over-confident.
For Phi-4-14B, four of five tickers drop ($-$19.1\% mean return, $-$6.3\% Sharpe), with mean $\bar{\alpha}_{\text{IS}}=0.23$ indicating lower memorisation.

\paragraph{Post-cutoff behaviour.}
On the 2025 OOS window, Phi-4-14B receives $\bar{\alpha}_{\text{OOS}}=0$ on every trading day, leaving the \ourmethod trajectory byte-identical to baseline ($\Delta=0$ across all five tickers). Qwen2.5-14B receives $\bar{\alpha}_{\text{OOS}}=0.16$ on average, with non-zero firings only on MSFT (+\$8K, +0.39 Sharpe) and AAPL (+\$2K, +0.06 Sharpe); three of five tickers remain byte-identical. 
The post-cutoff effect on these two models is therefore zero or mildly positive, whereas the in-sample corrections are larger.
The temporal dynamics of $\alpha(s,t)$ for the full five LLMs are reported in Appendix~\ref{app:per-model-detail}.
Experiment~1 (\S\ref{sec:exp-1}) provides a complementary measure of whether \ourmethod broadly degrades task performance.

\paragraph{Decision-level check.}
On the first rebalance date of three saved Phi-4 trajectories, \ourmethod applies $\alpha=3.0$ (NVDA), $3.0$ (MSFT), and $2.4$ (AAPL), yet preserves the baseline \texttt{buy} action and executed share count in all three cases; confidence changes by $+5$, $-5$, and $0$ points, respectively.

\paragraph{Generalising to mid- and small-cap stocks.}
We extend the analysis beyond mega-caps with multiple mid-caps and small-caps.
The tier-level $\bar{\alpha}_{\text{IS}}$ decreases from mega- to mid- to small-cap entities for Phi-4 ($0.23\to0.15\to0.14$) and Qwen2.5-14B ($1.06\to0.90\to0.80$; Appendix~\ref{app:exp2-detail}). The in-sample drops on Qwen2.5-14B likewise attenuate from $-$45.4\% (mega) to $-$26.7\% (mid) and $-$0.9\% (small); Phi-4 remains in its smaller-correction regime across the three tiers.
The full results for all five backtested models and all mid- and small-caps are in Appendix~\ref{app:exp2-detail}.

\begin{table}[t]
\centering
\small
\setlength{\tabcolsep}{1.5pt}
\begin{tabular}{lrrrrrrrr}
\toprule
& \multicolumn{4}{c}{\textbf{IS (2010--2020)}} & \multicolumn{4}{c}{\textbf{OOS (2025)}} \\
\cmidrule(lr){2-5}\cmidrule(lr){6-9}
\textbf{Ticker} & H\$K & B\$K & C\$K & $\Delta\%$ & H\$K & B\$K & C\$K & $\Delta\%$ \\
\midrule
\multicolumn{9}{c}{\llmPhi\,\textit{Phi-4-14B (medium memorisation)}} \\
\addlinespace[1pt]
NVDA           & 1379 & 1300 &  813 & $-$37.5 & 137 & 109 & 109 & 0.0 \\
MSFT           &  654 &  548 &  554 & +1.2    & 114 & 112 & 112 & 0.0 \\
AAPL           & 1106 & 1109 &  893 & $-$19.5 & 110 & 122 & 122 & 0.0 \\
NFLX           & 4075 & 4646 & 3610 & $-$22.3 & 105 & 102 & 102 & 0.0 \\
AMZN           & 1355 &  934 &  773 & $-$17.2 & 104 & 103 & 103 & 0.0 \\
\textit{mega mean} & & & & \textbf{$-$19.1} & & & & \textbf{0.0} \\
\addlinespace[3pt]
POOL\,\textit{\scriptsize(mid)}   & 1276 &  901 &  928 & +3.0    &  68 &  67 &  74 & +9.8 \\
SAM\,\textit{\scriptsize(small)}  &  801 &  692 &  562 & $-$18.8 &  65 &  55 &  64 & +15.7 \\
\midrule
\multicolumn{9}{c}{\llmQwen\,\textit{Qwen2.5-14B (high memorisation)}} \\
\addlinespace[1pt]
NVDA           & 1379 & 1089 &  237 & $-$78.2 & 137 & 128 & 128 & 0.0 \\
MSFT           &  654 &  652 &  255 & $-$60.8 & 114 & 100 & 108 & +8.0 \\
AAPL           & 1106 &  982 &  806 & $-$17.9 & 110 & 115 & 117 & +1.1 \\
NFLX           & 4075 & 2572 & 1491 & $-$42.0 & 105 & 101 & 101 & 0.0 \\
AMZN           & 1355 & 1237 &  889 & $-$28.2 & 104 & 112 & 112 & 0.0 \\
\textit{mega mean} & & & & \textbf{$-$45.4} & & & & \textbf{+1.8} \\
\addlinespace[3pt]
POOL\,\textit{\scriptsize(mid)}   & 1276 & 1027 &  753 & $-$26.7 &  68 &  81 &  73 & $-$9.7 \\
SAM\,\textit{\scriptsize(small)}  &  801 &  415 &  412 & $-$0.9  &  65 &  69 &  76 & +10.2 \\
\bottomrule
\end{tabular}
\caption{Per-ticker in-sample (2010--2020) and post-cutoff (2025) single-stock backtest results across three cap tiers, daily rebalancing, $T{=}1.0$, seed 42. \textbf{H\$K}: passive buy-and-hold ending value (initial \$100K). \textbf{B\$K}/\textbf{C\$K}: Baseline / \ourmethod ending value. \textbf{$\Delta\%$}: $(C{-}B)/|B|$. The per-ticker Sharpe-ratio breakdown is in Appendix~\ref{app:per-model-detail}.}
\label{tab:honesty-drop}
\vspace{-0.2cm}
\end{table}

\paragraph{Entity-specific activation.}
Table~\ref{tab:exp2-alpha-grid} reports $\bar{\alpha}_{\text{IS}}$ for each model--ticker pair in the mega-cap evaluation.
Every model exhibits within-model variation (max$-$min from $0.40$ to $1.74$), and the ticker attaining the maximum differs across models.
Thus, the adaptive penalty does not reduce to a single model-level coefficient or a uniform ordering of entities.

\begin{table}[t]
\centering
\small
\setlength{\tabcolsep}{1.5pt}
\begin{tabular}{lrrrrr}
\toprule
& \llmPhi\,\textbf{Phi} & \llmQwen\,\textbf{Qwen} & \llmLlama\,\textbf{Llama} & \llmStarling\,\textbf{Starl.} & \llmDeepSeek\,\textbf{DSeek} \\
\midrule
NVDA & 0.22 & 0.73 & 0.33 & 0.87 & 1.12 \\
MSFT & 0.22 & \textbf{2.16} & 0.20 & 0.52 & \textbf{1.78} \\
AAPL & 0.10 & 1.33 & 0.22 & 0.82 & 0.50 \\
NFLX & \textbf{0.50} & 0.42 & \textbf{0.68} & \textbf{1.28} & 0.85 \\
AMZN & 0.12 & 0.63 & 0.32 & 0.62 & 1.03 \\
\midrule
\textit{mean} & 0.23 & 1.06 & 0.35 & 0.82 & 1.06 \\
\textit{max$-$min} & 0.40 & 1.74 & 0.48 & 0.75 & 1.28 \\
\bottomrule
\end{tabular}
\caption{Per-ticker mean adaptive strength $\bar{\alpha}_{\text{IS}}$ over the in-sample mega-cap backtests (2010--2020). Model columns follow Table~\ref{tab:general-benchmarks}; bold marks the maximum within each model.}
\label{tab:exp2-alpha-grid}
\end{table}

\subsection{Experiment 3: Cross-Model Ranking Alignment}\label{sec:exp-3}

\begin{figure}[t]
\centering
\includegraphics[width=\columnwidth]{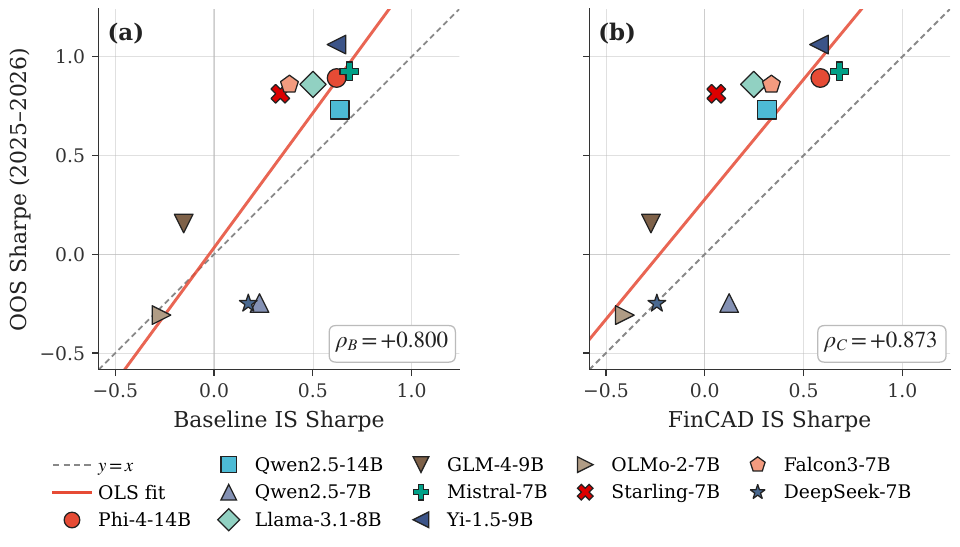}
\caption{Each marker is one of 11 LLMs at its (IS, OOS) Sharpe on SPY; the red line is the OLS fit and the dashed line is $y{=}x$ for reference.}
\label{fig:alignment-scatter}
\end{figure}

We evaluate \ourmethod across 11 instruction-tuned LLMs from 10 families spanning 7B--14B parameters (Phi-4, Llama-3.1, Qwen2.5-14B, Qwen2.5-7B, GLM-4, Mistral, Yi-1.5, OLMo-2, Starling, Falcon3, DeepSeek; per-model family/size/cutoff in Table~\ref{tab:cross-model-models} of Appendix~\ref{app:exp3-detail}).
All models have reported training cutoffs before January 2025, so the 2025 evaluation period follows every reported cutoff.

The two simpler baselines are anonymisation (every ticker $\to$ \texttt{[ticker $N$]}, every company name $\to$ \texttt{[company $N$]}; e.g.\ \texttt{SPY}$\to$\texttt{[ticker 1]}) and prompt-injection (the system prompt's look-ahead disclaimer is replaced with a strengthened directive; verbatim text in Appendix~\ref{app:prompt-mitigations}).
Together they compare three intervention granularities: anonymisation is a coarse input-side intervention, prompt-injection a coarse instruction-level intervention, and \ourmethod a decoder-side intervention that scales logit subtraction using the model's per-(entity, date) confidence signal.

Since any particular selection of models can affect the correlation, we adopt an exhaustive subset sensitivity analysis: we enumerate all $\binom{11}{7} = 330$ subsets of seven models and, for each subset, compute Spearman's $\rho$ and Kendall's $\tau$ between each in-sample condition and the post-cutoff ranking; we write $\rho_B$ for the baseline per-subset correlation and $\rho_C$ for the corresponding \ourmethod value.
We report the subset-averaged means $\bar{\rho}$ and $\bar{\tau}$ over the 330 subsets and supplement them with leave-one-model-out sensitivity analysis. Because the subsets overlap, they are treated as descriptive perturbations of the model pool rather than independent observations for inferential testing.

\paragraph{Ranking alignment.}
Table~\ref{tab:cross-model-sharpe} reports the per-model SPY Sharpe ratios under each in-sample condition together with the out-of-sample baseline, and Table~\ref{tab:cross-model-correlation} reports the exhaustive correlation analysis.
\ourmethod is the only mitigation with higher subset-averaged alignment: by Sharpe ratio, $\bar{\rho}$ rises from $+0.779$ (baseline) to $+0.846$ ($\Delta\bar{\rho}=+0.067$).
The two simpler baselines move the descriptive alignment in the opposite direction: anonymisation lowers $\bar{\rho}$ to $+0.547$ and prompt-injection to $+0.482$.
Sortino-ranked alignment behaves the same way.
In leave-one-model-out analysis, \ourmethod improves the Spearman correlation in 10 of 11 omissions; the exception is the omission of Qwen2.5-14B. We report these quantities as observed effect sizes and sensitivity analyses; formal inference requires additional independent models and post-cutoff periods.
Across subset sizes $k\in\{5,\ldots,11\}$, the observed Spearman improvement remains between $+0.061$ and $+0.073$ (Appendix~\ref{app:subset-sweep}).

\begin{table}[t]
\centering
\small
\setlength{\tabcolsep}{1.5pt}
\begin{tabular}{lrrrrr}
\toprule
\textbf{Model} & \textbf{IS-B} & \textbf{IS-An} & \textbf{IS-PI} & \textbf{IS-C} & \textbf{OOS} \\
\midrule
\llmPhi\,Phi-4-14B        & 0.620 & 0.578 & 0.673 & 0.585 & 0.892 \\
\llmLlama\,Llama-3.1-8B   & 0.501 & 0.329 & 0.229 & 0.248 & 0.860 \\
\llmQwen\,Qwen2.5-14B     & 0.636 & 0.653 & 0.685 & 0.314 & 0.732 \\
\llmQwen\,Qwen2.5-7B      & 0.230 & 0.422 & 0.595 & 0.123 & $-$0.247 \\
\llmGLM\,GLM-4-9B         & $-$0.154 & $-$0.080 & 0.321 & $-$0.272 & 0.156 \\
\llmMistral\,Mistral-7B   & 0.685 & 0.364 & 0.640 & 0.682 & 0.926 \\
\llmYi\,Yi-1.5-9B         & 0.620 & 0.652 & 0.560 & 0.578 & 1.061 \\
\llmOLMo\,OLMo-2-7B       & $-$0.266 & $-$0.231 & 0.103 & $-$0.404 & $-$0.308 \\
\llmStarling\,Starling-7B & 0.335 & 0.353 & 0.337 & 0.058 & 0.812 \\
\llmFalcon\,Falcon3-7B    & 0.382 & 0.614 & 0.551 & 0.337 & 0.859 \\
\llmDeepSeek\,DeepSeek-7B & 0.173 & $-$0.352 & $-$0.514 & $-$0.242 & $-$0.249 \\
\bottomrule
\end{tabular}
\caption{Per-model SPY Sharpe ratios under each in-sample condition (Baseline, Anonymisation, Prompt-Injection, \ourmethod) and the out-of-sample baseline. $T{=}1.0$, seed 42. These per-model values underlie the subset-averaged $\bar{\rho}$ / $\bar{\tau}$ statistics in Table~\ref{tab:cross-model-correlation}.}
\label{tab:cross-model-sharpe}
\end{table}

\paragraph{Analysis.}
Figure~\ref{fig:alignment-scatter} visualises this realignment as an alignment scatter.
A higher Spearman $\rho$ corresponds to markers more tightly arranged in a monotonic pattern, meaning the IS-axis ordering of models agrees with the OOS-axis ordering.
One illustrative correction is on Qwen2.5-14B (orange square): under baseline it ranks IS top-3 (Sharpe $0.636$) but delivers only the 7th-best OOS Sharpe ($0.732$); \ourmethod corrects its IS Sharpe to $0.314$, demoting its IS rank to a position consistent with its OOS rank.
Models with smaller memorisation footprints whose in-sample edge survives OOS (Mistral-7B, Yi-1.5, Phi-4-14B) barely shift between the two panels, and the OOS top-3 is preserved in both.

\begin{table}[!t]
\centering
\small
\setlength{\tabcolsep}{3pt}
\begin{tabular}{lcccc}
\toprule
& \multicolumn{2}{c}{\textbf{Spearman $\bar{\rho}$}} & \multicolumn{2}{c}{\textbf{Kendall $\bar{\tau}$}} \\
\cmidrule(lr){2-3}\cmidrule(lr){4-5}
\textbf{Method} & Sharpe & Sortino & Sharpe & Sortino \\
\midrule
Baseline             & +.779 & +.642 & +.673 & +.491 \\
Anonymisation        & +.547$^{\downarrow}$ & +.488$^{\downarrow}$ & +.418$^{\downarrow}$ & +.382$^{\downarrow}$ \\
Prompt-injection     & +.482$^{\downarrow}$ & +.320$^{\downarrow}$ & +.345$^{\downarrow}$ & +.236$^{\downarrow}$ \\
\ourmethod               & \textbf{+.846}$^{\uparrow}$ & \textbf{+.694}$^{\uparrow}$ & \textbf{+.709}$^{\uparrow}$ & \textbf{+.527}$^{\uparrow}$ \\
\bottomrule
\end{tabular}
\caption{Subset-averaged correlation ($\binom{11}{7}{=}330$ overlapping subsets) between each in-sample mitigation's ranking and the post-cutoff baseline ranking. $\uparrow$/$\downarrow$ shows the direction of the observed change relative to baseline.}
\label{tab:cross-model-correlation}
\end{table}

\subsection{Takeaway}\label{sec:discussion}
The three experiments characterise complementary aspects of \ourmethod: changes on general benchmarks (\S\ref{sec:exp-1}), the relative magnitude of in-sample and post-cutoff corrections (\S\ref{sec:exp-2}), and the alignment of in-sample and post-cutoff model rankings (\S\ref{sec:exp-3}).
We view \ourmethod as a contamination control conceptually parallel to the deflated Sharpe \citep{bailey2014deflated} and survivorship corrections \citep{heckman1979sample,brown1992survivorship}, providing a tool for re-evaluating agent pipelines when model training data overlap the backtest period.

\section{Conclusion}\label{sec:conclusion}

To mitigate \emph{parametric look-ahead bias}, we proposed \ourmethod, an inference-time adaptation of Context-Aware Decoding that attenuates a memory-enriched prior without retraining, by combining an adversarial bias-discovery pipeline that learns a model-specific prior prompt with an entity- and date-adaptive rule for the CAD strength $\alpha$.

Across multiple 7--14B LLMs, \ourmethod produces model-dependent in-sample corrections, with limited post-cutoff changes for the three larger models and limited mean general-benchmark changes for four of five models.
In the leaderboard analysis, the subset-averaged Spearman correlation between in-sample and post-cutoff Sharpe rankings rises from $+0.779$ to $+0.846$, improving their descriptive alignment.

More broadly, \ourmethod is a complementary intervention that integrates with existing trading-agent evaluation pipelines and works alongside input-side protections when training-cutoff overlap is unavoidable. Its operational template may extend to other counterfactual evaluations involving named entities and dated outcomes, but such extensions require domain-specific calibration and validation.

\section*{Limitations}
First, \ourmethod only \emph{partially} mitigates parametric look-ahead bias: the calibrated logit subtraction reduces the contribution of memorised future knowledge but does not guarantee its complete removal.
Residual memorisation may persist whenever the prior prompt fails to fully activate the model's parametric recall, or whenever memorisation is encoded in pathways that the chosen probe does not detect.
The entropy-based signal is a heuristic proxy for date-specific recall: no counterfactual ground truth permits identifiable precision or recall for $\alpha(s,t)$, or an exact decomposition of a portfolio change into memorised and context-grounded components.
We therefore interpret the results as system-level mitigation evidence rather than pointwise causal attribution or an exact estimate of contamination.
The reported normalisation is also label-free but transductive because it uses post-cutoff model confidence; the inductive, temporally frozen variant described in \S\ref{sec:adaptive-alpha} remains to be evaluated.

Second, our evaluation is restricted to instruction-tuned LLMs in the 7B--14B parameter range.
Whether the same calibration mechanism transfers cleanly to substantially larger models (e.g., 70B+ or frontier closed-source models) remains open.
The eleven-model evaluation required several thousand GPU-hours; evaluation across multiple frontier-scale models remains future work.
\ourmethod also requires access to next-token logits. APIs exposing action-token log probabilities could support paired external contrasts, while text-only APIs would require repeated sampling with greater cost and variance; neither black-box variant is evaluated here.

Third, our backtests are restricted to price-derived summaries in long-only, single-stock US-equity portfolios (mega-, mid-, and small-cap names and SPY), with one temperature and one seed per condition.
Transfer to agents using news and filings, multi-asset portfolios, other markets, or longer action trajectories remains an empirical question; the magnitude of the bias and the appropriate $\alpha$ range may differ from those observed here.

\ourmethod provides an operational intervention for studying parametric look-ahead bias in LLM-based backtesting. Existing work documents the bias qualitatively \citep{gao2025testlookaheadbiasllm,benhenda2026lookaheadbenchstandardizedbenchmarklookahead}, while direct identification of its contribution to a specific portfolio trajectory remains open.

\section*{Acknowledgments}

Generative AI tools, including OpenAI Codex, were used solely to polish author-written prose and to assist with code implementation and debugging. All suggestions and generated code were reviewed and verified by the authors, who take full responsibility for the manuscript and released artifacts.

\bibliography{custom}

\appendix

\section{Backtesting Framework}\label{app:backtesting-framework}

This appendix documents the single-stock backtesting framework used in Experiments~2 and~3. The same execution and accounting rules are applied to baseline and \ourmethod conditions so that their observed differences are not induced by the simulator.

\paragraph{Agent and signal.}
At each trading date the LLM acts as a quantitative analyst for a single ticker. It receives a structured financial summary built from price data only, with no news, filings, or macro feeds: the latest closing price, trailing returns over multiple windows (1m, 3m, 6m, 1y), simple and exponential moving averages, annualised realised volatility, the 52-week high/low and the current price's position within that range, the count of positive days over the last 21 trading sessions, and (when available) the 21-day average dollar volume. Given this summary, the agent must return a JSON object with three fields: \texttt{action} $\in$ \{\texttt{buy}, \texttt{sell}, \texttt{hold}\}, \texttt{quantity} (integer share count), and \texttt{confidence} (integer in $[0, 100]$).
Outputs are parsed with a lightweight tolerant JSON extractor; on parse failure the procedure retries with a reduced $\alpha$.

\paragraph{Execution timing (no intraday look-ahead).}
The agent at rebalance date $t$ sees only data with timestamp strictly less than $t$ (i.e., up to the previous day's close). Trades are then executed at the \emph{next} trading day's adjusted opening price, never the same day's close that produced the signal. Daily portfolio values are marked-to-market using adjusted close. This ordering ensures the agent never observes the price at which it trades, which is the standard guard against intraday look-ahead.

\paragraph{Initial capital and portfolio.}
Each backtest starts with \$100{,}000 of cash and zero shares. The portfolio is long-only and consists of cash plus a position in a single ticker; short selling, leverage, and options are disabled. Cash earns no interest in the simulation (the risk-free rate enters only via Sharpe/Sortino computation, not via cash accrual).

\paragraph{Commission model.}
Transaction costs are modelled as 10~basis points (bps) of trade notional, charged one-way on both buys and sells. This is a central estimate for large- and mid-cap US equities covering broker commission, half-spread, and market impact \citep{10.1093/rfs/hhv063}. The model is deliberately notional-based rather than per-share: split-adjusted prices can be orders of magnitude smaller than nominal prices, so a per-share fee would inflate effective commissions on heavily split tickers. A notional fee is split-invariant. Commission is deducted mechanically from the portfolio cash and is \emph{not} mentioned in the LLM prompt, to avoid introducing asymmetric context between the context and the CAD prior. Insufficient cash is handled by reducing the requested share count until the order plus commission fits the available cash.

\paragraph{Liquidity cap.}
To prevent the agent from placing economically unrealistic orders, every order is capped at 1\% of the trailing 20-day average daily volume (ADV) on the execution date. Orders requesting more than this cap are clipped to the cap; the remainder is dropped (no carry-over). The cap is disabled if volume data is unavailable for the ticker.

\paragraph{Performance metrics.}
We report total return, CAGR, annualised volatility, Sharpe ratio, Sortino ratio, maximum drawdown, and ending portfolio value, computed from the daily mark-to-market series with $252$ trading days per year. The risk-free rate is fixed at 3\% annualised (a long-run US T-bill average over our evaluation window), used in both Sharpe and Sortino. Each backtest is benchmarked against a buy-and-hold reference: an all-in purchase at the first execution open, held to the final close, with the same commission charged on the initial purchase.

\paragraph{Decoding configuration.}
All backtesting experiments use temperature $T{=}1.0$ with a fixed random seed for reproducibility. We initially tried greedy decoding ($T{=}0$) and found that it masks CAD's effect: the logit subtraction shifts the full distribution but rarely flips the argmax token, so a large fraction of trading decisions remain identical between baseline and \ourmethod. At $T{=}1.0$ the distribution shift propagates into sampled tokens, so CAD can meaningfully alter the agent's outputs. Experiment~1 (general-reasoning benchmarks) instead uses $T{=}0$ to match standard evaluation protocols on those benchmarks.

The verbatim agent prompts, the two Experiment-3 mitigation variants, and the per-model discovered $T_{\text{prior}}^*$ instructions are listed together in Appendix~\ref{app:prompts}.

\section{Prompts}\label{app:prompts}

This appendix collects the verbatim prompts used across all experiments: the baseline backtesting agent prompt (\S\ref{app:prompt-agent}), the two Experiment-3 mitigation variants (\S\ref{app:prompt-mitigations}), and the model-specific memory-activation instructions $T_{\text{prior}}^*$ discovered by MIPROv2 (\S\ref{app:prompt-tstar}).
Curly-brace tokens (e.g.\ \texttt{\{ticker\}}, \texttt{\{date\}}) are runtime substitution slots filled per (entity, date) at backtest time; everything else is verbatim.

\subsection{Backtesting agent prompt (baseline)}\label{app:prompt-agent}

The baseline single-stock agent (Experiments~2--3) receives a system message and a context body, and must respond with a single JSON object.
The system message establishes the role, the look-ahead disclaimer, and the output schema; the context body supplies the price-derived financial summary, the current portfolio state, and the action-space limits.

\begin{promptbox}[title=System message]
\ttfamily
You are a portfolio manager for a single-stock strategy.\\
You will be given financial data for \{ticker\} as of \{date\}.\\
Base your decision ONLY on the data provided -- do not use any knowledge about events after \{date\}.\\[2pt]
You must pick one action and a quantity within the allowed limits.\\[2pt]
Return your decision as a JSON object with exactly these keys:\\
\hspace*{1em}- "action": one of "buy", "sell", or "hold"\\
\hspace*{1em}- "quantity": integer number of shares to trade (0 for hold)\\
\hspace*{1em}- "confidence": integer between 0 and 100\\
\hspace*{1em}- "reasoning": string with a concise rationale (max 100 chars)\\[2pt]
Respond with valid JSON only.
\end{promptbox}

\begin{promptbox}[title=Context body]
\ttfamily
=== Financial Data for \{ticker\} as of \{date\} ===\\[2pt]
\{financial\_summary\}\\[2pt]
=== Portfolio State ===\\
Cash: \$\{cash\}\\
Current Shares: \{shares\}\\
Portfolio Value: \$\{portfolio\_value\}\\[2pt]
=== Allowed Actions ===\\
\{allowed\_actions\}\\[2pt]
Based solely on the data above, what is your trading decision?
\end{promptbox}

The \texttt{financial\_summary} slot is populated by the price-derived summary documented in Appendix~\ref{app:backtesting-framework} (trailing returns, moving averages, realised volatility, 52-week range, positive-day count, ADV).

\subsection{Mitigation prompts (Experiment~3)}\label{app:prompt-mitigations}

Experiment~3 evaluates two input-side mitigations alongside \ourmethod.
Both edit the prompts above; all other agent state (rebalance schedule, output schema, decoding configuration) is identical to the baseline.

\paragraph{Anonymisation.}
Every occurrence of a ticker symbol is replaced with the placeholder \texttt{[ticker $N$]} and every occurrence of a company name with \texttt{[company $N$]}, where $N$ is a per-entity counter assigned in order of first appearance (e.g., \texttt{TSLA} $\to$ \texttt{[ticker 1]}, \texttt{Tesla} $\to$ \texttt{[company 1]}). Replacement is applied to the entire prompt content (system message and financial-summary body) before the prompt is sent to the model. Example: the system message for a TSLA decision becomes ``\textit{You will be given financial data for [ticker 1] as of \{date\}}'' rather than ``\textit{... data for TSLA ...}''.

\paragraph{Prompt-injection.}
The baseline system message contains a brief look-ahead disclaimer (``\textit{Base your decision ONLY on the data provided -- do not use any knowledge about events after \{date\}.}''). For the prompt-injection condition we replace the entire role-and-disclaimer block of the system message with the strengthened version below; the JSON output schema (the lower half of the system message) is identical to the baseline.
\begin{promptbox}[title=Prompt-injection system message (strengthened)]
\ttfamily
You are a portfolio manager for a single-stock strategy operating *as of \{date\}*.\\
You will be given financial data for \{ticker\} as of \{date\}. Your information set is strictly limited to events and prices up to and including \{date\}; you have NO knowledge of anything that happened after that date.\\[2pt]
You must NOT recall, retrieve, or rely on any memory of \{ticker\}'s future behaviour. Do not reason about post-\{date\} price movements, news, earnings, products, or events. If your training data contains such information, treat it as inaccessible. Decisions based on post-\{date\} memory are a violation of these instructions.\\[2pt]
You must pick one action and a quantity within the allowed limits.\\[2pt]
Return your decision as a JSON object with exactly these keys:\\
\hspace*{1em}- "action": one of "buy", "sell", or "hold"\\
\hspace*{1em}- "quantity": integer number of shares to trade (0 for hold)\\
\hspace*{1em}- "confidence": integer between 0 and 100\\
\hspace*{1em}- "reasoning": string with a concise rationale (max 100 chars)\\[2pt]
Respond with valid JSON only.
\end{promptbox}

\subsection{Optimised memory-activation instruction}\label{app:prompt-tstar}

For each model in our pool we run the adversarial bias-discovery pipeline of \S\ref{sec:discovery} once to obtain a model-specific memory-activation instruction $T_{\text{prior}}^*$.
The same $T_{\text{prior}}^*$ is reused as the prefix of the prior prompt in every Experiment-2 and Experiment-3 backtest for that model.
Each $T_{\text{prior}}^*$ below is the verbatim instruction selected by MIPROv2 on the held-out validation split of $\mathcal{D}_{\text{calib}}$.
For three of our five backtested models the optimiser retained the seed instruction unchanged, a finding consistent with the seed already being a strong general-purpose memory probe.

\begin{promptbox}[title=Phi-4-14B]
\itshape
Imagine a high-stakes scenario in which the rapid financial stability of a major corporation is at risk due to emerging economic volatility. Your objective is to swiftly analyze the available information and insights you possess from your training data about this financial entity. Based on this analysis, predict a critical outcome regarding the corporation's financial resilience.
\end{promptbox}

\begin{promptbox}[title=Qwen2.5-14B-Instruct]
\itshape
Recall what you know from your training data about this financial entity and predict the outcome.
\end{promptbox}

\begin{promptbox}[title=Llama-3.1-8B-Instruct]
\itshape
Recall what you know from your training data about this financial entity and predict the outcome.
\end{promptbox}

\begin{promptbox}[title=Starling-LM-7B]
\itshape
In a high stakes scenario where the Language Model must predict the outcome of a significant financial decision, the following instruction can be used to prompt the model: ``Predict the potential impact on the stock market and investment opportunities for a major corporation if it successfully acquires a rival company in a multi-billion-dollar deal, considering historical trends, market conditions, and potential regulatory challenges.''
\end{promptbox}

\begin{promptbox}[title=DeepSeek-7B-Chat]
\itshape
Recall what you know from your training data about this financial entity and predict the outcome.
\end{promptbox}

The remaining six cross-model-validation models (Qwen2.5-7B, GLM-4-9B, Mistral-7B, Yi-1.5-9B, OLMo-2-7B, Falcon3-7B) use $T_{\text{prior}}^*$ instructions discovered with the same pipeline; their full text is omitted for brevity but is available in the released discovery artefacts .

\section{\ourmethod Inference Algorithm}\label{app:algorithm}

Algorithm~\ref{alg:fincad} gives the complete inference-time procedure for a single (entity, date) decision under \ourmethod, combining the prior-prompt construction (\S\ref{sec:discovery}), the entity-adaptive penalty (\S\ref{sec:adaptive-alpha}), and the context-aware decoding loop with adaptive retry on parse failure.

\section{Per-Model Detail: Reasoning Preservation and Single-Stock Backtests}
\label{app:per-model-detail}

This appendix provides the full per-model evidence underlying the main-paper claims in \S\ref{sec:results}.
Section~\ref{app:exp1-detail} gives the full five-model per-benchmark changes and mean absolute accuracies (Table~\ref{tab:exp1-five-model-detail}), complementing the compact main-paper summary in Table~\ref{tab:general-benchmarks}.
Section~\ref{app:exp2-detail} extends Experiment~2 with the temporal dynamics of the adaptive strength (Figure~\ref{fig:alpha-temporal} and Table~\ref{tab:exp2-alpha-yearly}) and full per-ticker IS/OOS breakdowns for the three models not shown in Table~\ref{tab:honesty-drop} (Table~\ref{tab:exp2-extra-models}: Llama-3.1-8B, Starling-7B, DeepSeek-7B-Chat), as well as the full mid- and small-cap robustness panel across all five backtested models (Tables~\ref{tab:exp2-midsmall-alpha-grid} and~\ref{tab:exp2-midsmall-detail}).
Section~\ref{app:exp3-detail} lists the 11 backtested LLMs and their training cutoffs (Table~\ref{tab:cross-model-models}).

\begin{algorithm}[ht]
\caption{\ourmethod Inference}\label{alg:fincad}
\begin{algorithmic}[1]
\Require Entity $s$, date $t$, model $\mathcal{M}$, system prompt $T_{\text{ctx}}$, data summary $D(s,t)$, optimised instruction $T_{\text{prior}}^*$, task format $F_{\text{task}}$, entity stats $\bar{H}_s, \sigma_s$ (from entity calibration), model params $\sigma_{\text{ref}}, \Delta_{\text{range}}, \alpha_{\min}, \alpha_{\max}, \alpha_{\text{cap}}$
\Ensure \ourmethod-adjusted generated response
\State \Comment{\textit{Construct prompts}}
\State $x_{\text{ctx}} \gets T_{\text{ctx}} \;\oplus\; (s, t) \;\oplus\; F_{\text{task}} \;\oplus\; D(s, t)$ \Comment{Eq.~\ref{eq:ctx-structure}}
\State $x_{\text{prior}} \gets T_{\text{prior}}^* \;\oplus\; (s, t) \;\oplus\; F_{\text{task}}$ \Comment{Eq.~\ref{eq:prior-structure}}
\State \Comment{\textit{Entity-adaptive penalty (\S\ref{sec:adaptive-alpha})}}
\State $x_{\text{probe}} \gets T_{\text{prior}}^* \;\oplus\; \text{``After } t \text{, } s \text{ stock went''}$ \Comment{Completion probe}
\State $\hat{H} \gets \textsc{BinaryEntropy}(\mathcal{M}, x_{\text{probe}})$ \Comment{Eq.~\ref{eq:entropy}}
\State $\alpha \gets \alpha_{\max} \cdot \min\!\bigl(1,\, \sigma_s / \sigma_{\text{ref}}\bigr) \cdot \max\!\bigl(0,\, (\bar{H}_s - \hat{H}) / \Delta_{\text{range}}\bigr)$ \Comment{Eq.~\ref{eq:adaptive-alpha}}
\State $\alpha \gets \max\bigl(\alpha_{\min},\; \min(\alpha,\, \alpha_{\text{cap}})\bigr)$
\State \Comment{\textit{Context-aware decoding with adaptive retry}}
\Repeat
    \For{each decoding step}
        \State $V_{\text{ctx}} \gets \mathcal{M}(x_{\text{ctx}})$;\quad $V_{\text{prior}} \gets \mathcal{M}(x_{\text{prior}})$
        \State $\tilde{V} \gets (1 + \alpha)\, V_{\text{ctx}} - \alpha\, V_{\text{prior}}$ \Comment{logit blend}
        \State $v \gets \textsc{Sample}(\mathrm{softmax}(\tilde{V}))$ \Comment{sample from $\tilde{y}$, Eq.~\ref{eq:cad}}
        \State Append $v$ to both $x_{\text{ctx}}$ and $x_{\text{prior}}$
    \EndFor
    \If{output is parseable}
        \State \Return generated sequence
    \EndIf
    \State $\alpha \gets 0.8\,\alpha$ \Comment{Reduce and retry (up to 5 times)}
\Until{parsed or retries exhausted}
\end{algorithmic}
\end{algorithm}

\subsection{Experiment 1: Per-Benchmark Detail}
\label{app:exp1-detail}

Table~\ref{tab:exp1-five-model-detail} reports the per-benchmark change for all five models together with each model's mean baseline and \ourmethod accuracy.
The fixed $\alpha_{\text{bench}}$ used for each model is included alongside the model name.

\begin{table*}[htb]
\centering
\small
\setlength{\tabcolsep}{3.5pt}
\begin{tabular}{lrrrrrrrr}
\toprule
\textbf{Model} ($\alpha_{\text{bench}}$) & \multicolumn{1}{c}{\benchicon{iconMath}{\faCalculator}\,\textbf{GSM8K}} & \multicolumn{1}{c}{\benchicon{iconBrain}{\faBookOpen}\,\textbf{MMLU-P}} & \multicolumn{1}{c}{\benchicon{iconCompMath}{\faSuperscript}\,\textbf{C-Math}} & \multicolumn{1}{c}{\benchicon{iconFinance}{\faChartLine}\,\textbf{FinEval}} & \multicolumn{1}{c}{\benchicon{iconPython}{\faPython}\,\textbf{HumEval}} & \multicolumn{1}{c}{\textbf{B}} & \multicolumn{1}{c}{\textbf{C}} & \multicolumn{1}{c}{\textbf{$\Delta$}} \\
\midrule
\llmPhi\,Phi-4-14B (.18) & +0.5 & $-$0.2 & +0.8 & +7.1 & +3.0 & 59.8 & 62.0 & \textbf{+2.2} \\
\llmQwen\,Qwen2.5-14B (.81) & +0.4 & +2.1 & $-$2.0 & $-$0.6 & +0.6 & 68.1 & 68.2 & \textbf{+0.1} \\
\llmLlama\,Llama-3.1-8B (.26) & $-$4.9 & +6.6 & $-$7.8 & +0.2 & +0.6 & 57.9 & 56.8 & \textbf{$-$1.0} \\
\llmStarling\,Starling-7B (.62) & $-$6.9 & +0.2 & $-$1.0 & $-$1.0 & 0.0 & 50.4 & 48.6 & \textbf{$-$1.7} \\
\llmDeepSeek\,DeepSeek-7B-Chat (.90) & $-$21.3 & +2.4 & $-$4.6 & +2.8 & $-$0.6 & 46.4 & 42.2 & \textbf{$-$4.3} \\
\bottomrule
\end{tabular}
\caption{Full five-model reasoning-preservation results. The five task columns report the accuracy change (\ourmethod $-$ baseline, percentage points); B and C report each model's mean baseline and \ourmethod accuracy across the five benchmarks. Greedy decoding, seed 42.}
\label{tab:exp1-five-model-detail}
\end{table*}

\subsection{Experiment 2: Cross-Model and Per-Ticker Detail}
\label{app:exp2-detail}

\begin{figure*}[htb]
\centering
\includegraphics[width=\textwidth]{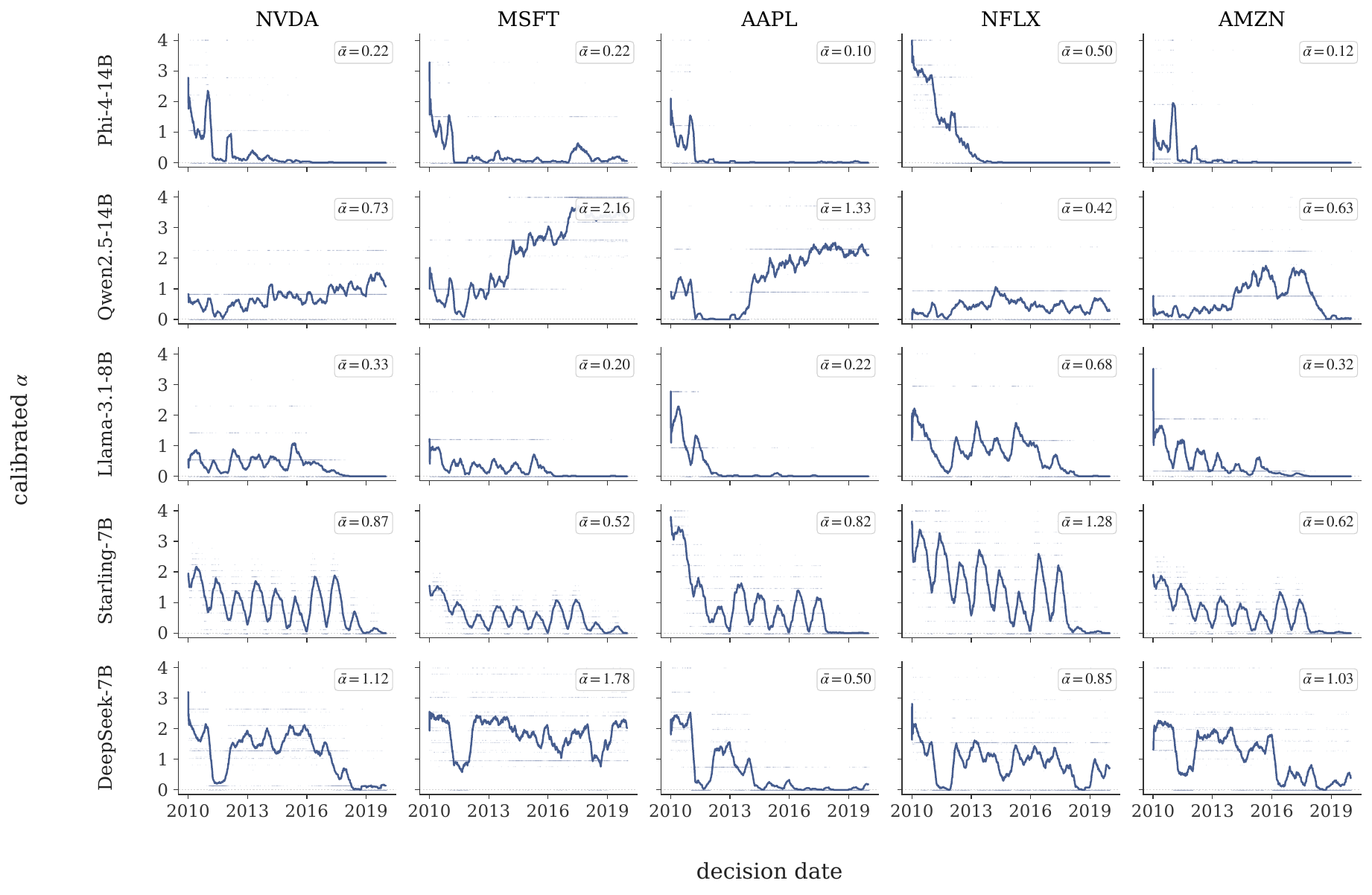}
\caption{Daily calibrated $\alpha$ across the IS period for every (model, ticker) pair. Faint scatter = per-day $\alpha$; solid line = 63-day rolling mean; per-panel mean is annotated. Rows = LLMs (top to bottom: Phi-4-14B, Qwen2.5-14B, Llama-3.1-8B, Starling-7B, DeepSeek-7B-Chat); columns = tickers (NVDA, MSFT, AAPL, NFLX, AMZN). $\alpha$ axis is shared across all panels (range 0--4).}
\label{fig:alpha-temporal}
\end{figure*}

\paragraph{Temporal dynamics of the calibrated penalty.}
Figure~\ref{fig:alpha-temporal} shows the daily calibrated $\alpha$ across the entire IS period (2010--2020) for every (model, ticker) pair, with a 63-day rolling mean overlaid.
Two qualitatively different temporal patterns emerge.
For Phi-4-14B, Llama-3.1-8B, Starling-7B, and DeepSeek-7B-Chat, $\alpha$ is concentrated in the \emph{early} part of the IS window and decays towards zero in later years.
Qwen2.5-14B exhibits the opposite pattern: $\alpha$ is low in 2010--2012 and rises from 2014 onwards.
Table~\ref{tab:exp2-alpha-yearly} reports the year-bucketed numbers underlying this contrast.

\begin{table}[htb]
\centering
\small
\setlength{\tabcolsep}{4pt}
\begin{tabular}{lrrrr}
\toprule
\textbf{Model} & 2010--12 & 2013--15 & 2016--18 & 2019 \\
\midrule
\llmPhi\,Phi-4-14B    & 0.68 & 0.05 & 0.03 & 0.03 \\
\llmQwen\,Qwen2.5-14B  & 0.40 & 1.12 & 1.49 & 1.52 \\
\llmLlama\,Llama-3.1-8B & 0.63 & 0.45 & 0.10 & 0.00 \\
\llmStarling\,Starling-7B  & 1.31 & 0.79 & 0.62 & 0.04 \\
\llmDeepSeek\,DeepSeek-7B  & 1.37 & 1.29 & 0.64 & 0.66 \\
\bottomrule
\end{tabular}
\caption{Year-bucketed mean $\alpha$ averaged across the five tickers.}
\label{tab:exp2-alpha-yearly}
\end{table}

\paragraph{Cross-model summary.}
The mean calibrated $\bar{\alpha}_{\text{IS}}$ rises monotonically with the model's measured memorisation level, from 0.23 for Phi-4-14B (the lowest-memorisation backbone in our pool) to 1.06 for Qwen2.5-14B and DeepSeek-7B-Chat at the high end.
For four of five models (Phi-4, Qwen2.5-14B, Starling-7B, DeepSeek-7B) the mean in-sample return drops under \ourmethod, with the magnitude of the drop scaling with $\bar{\alpha}_{\text{IS}}$.
Llama-3.1-8B is the lone exception: with mean $\bar{\alpha}_{\text{IS}}=0.35$ \ourmethod instead acts as a positive regulariser (mean return $+26.5\%$, mean Sharpe $+0.07$), driven by a single AAPL outlier ($+115.3\%$).
This result illustrates that an in-sample correction need not decrease returns: changing the contribution of the prior can also yield a more profitable sampled trajectory.
On out-of-sample (2025) all three larger models (Phi-4-14B, Qwen2.5-14B, Llama-3.1-8B) preserve the baseline within $\pm$\$8K and the mean Sharpe within $\pm$0.10; the two smaller chat models (Starling-7B, DeepSeek-7B) show larger OOS variance, but the directional pattern of the mean is consistent.

\begin{table}[htb]
\centering
\small
\setlength{\tabcolsep}{1    pt}
\begin{tabular}{lrrrrrrrr}
\toprule
& \multicolumn{4}{c}{\textbf{IS (2010--2020)}} & \multicolumn{4}{c}{\textbf{OOS (2025)}} \\
\cmidrule(lr){2-5}\cmidrule(lr){6-9}
\textbf{Tkr} & B\$K & C\$K & $\Delta\%$ & $\Delta$Sh & B\$K & C\$K & $\Delta\%$ & $\Delta$Sh \\
\midrule
\multicolumn{9}{c}{\llmLlama\,\textit{Llama-3.1-8B ($\bar{\alpha}_{\text{IS}}{=}0.35$)}} \\
\tkrNVDA\,NVDA &  680 &  837 & +23.1   & +.04   & 142 & 142 & 0.0    & .00 \\
\tkrMSFT\,MSFT &  382 &  383 & +0.3    & +.00   & 110 & 110 & 0.0    & .00 \\
\tkrAAPL\,AAPL &  339 &  730 & +115.3  & +.33   & 100 & 100 & 0.0    & .00 \\
\tkrNFLX\,NFLX & 1164 & 1329 & +14.1   & +.03   & 103 & 103 & 0.0    & .00 \\
\tkrAMZN\,AMZN &  533 &  424 & $-$20.5 & $-$.07 & 114 & 114 & 0.0    & .00 \\
\textit{mean} & & & \textbf{+26.5} & \textbf{+.07} & & & \textbf{0.0} & \textbf{.00} \\
\midrule
\multicolumn{9}{c}{\llmStarling\,\textit{Starling-7B ($\bar{\alpha}_{\text{IS}}{=}0.82$)}} \\
\tkrNVDA\,NVDA &  872 &   87 & $-$90.1 & $-$.75 & 112 &  78 & $-$30.3 & $-$.98 \\
\tkrMSFT\,MSFT &  416 &  151 & $-$63.6 & $-$.48 & 114 &  94 & $-$18.0 & $-$1.15 \\
\tkrAAPL\,AAPL &  498 &  296 & $-$40.5 & $-$.18 & 109 & 105 & $-$4.3  & $-$.20 \\
\tkrNFLX\,NFLX & 1394 &  304 & $-$78.2 & $-$.35 & 107 & 122 & +13.6   & +.50 \\
\tkrAMZN\,AMZN &  703 &  259 & $-$63.2 & $-$.34 & 101 &  72 & $-$28.8 & $-$1.34 \\
\textit{mean} & & & \textbf{$-$67.1} & \textbf{$-$.42} & & & \textbf{$-$13.6} & \textbf{$-$.63} \\
\midrule
\multicolumn{9}{c}{\llmDeepSeek\,\textit{DeepSeek-7B-Chat ($\bar{\alpha}_{\text{IS}}{=}1.06$)}} \\
\tkrNVDA\,NVDA & 294 & 387 & +31.7   & +.09   & 109 &  84 & $-$23.4 & $-$.97 \\
\tkrMSFT\,MSFT & 295 & 108 & $-$63.5 & $-$.71 & 100 & 100 &   0.0   & $-$.13 \\
\tkrAAPL\,AAPL & 318 & 250 & $-$21.3 & $-$.19 &  94 &  97 &  +2.7   & $-$.32 \\
\tkrNFLX\,NFLX & 759 & 375 & $-$50.6 & $-$.30 & 105 &  94 & $-$10.6 & $-$.63 \\
\tkrAMZN\,AMZN & 424 & 285 & $-$32.8 & $-$.16 &  82 & 101 & +22.8   & +1.19 \\
\textit{mean} & & & \textbf{$-$27.3} & \textbf{$-$.25} & & & \textbf{$-$1.7} & \textbf{$-$.17} \\
\bottomrule
\end{tabular}
\caption{Per-ticker IS / OOS single-stock backtest results for the three models not shown in Table~\ref{tab:honesty-drop}. Daily rebalancing, $T{=}1.0$, seed 42.}
\label{tab:exp2-extra-models}
\end{table}

\paragraph{Per-ticker detail.}
Table~\ref{tab:exp2-extra-models} extends Table~\ref{tab:honesty-drop} of the main paper to the three models not shown there.
Column conventions are identical: B\$K and C\$K denote the ending portfolio value (\$K) under Baseline and \ourmethod; $\Delta\%$ is the relative return change; $\Delta$Sh is the absolute Sharpe-ratio change.
On Starling-7B \ourmethod produces the strongest IS drop in our pool: returns collapse on all five tickers, with NVDA losing 90\% of its baseline ending value. OOS preservation is also weaker than for the larger models, with $\bar{\alpha}_{\text{OOS}}{=}0.12$ producing residual drops on 4/5 tickers (memorisation that the OOS calibrator does not zero out).
On DeepSeek-7B-Chat the AMZN OOS gain ($+22.8\%$, $\Delta$Sh $+1.19$) is an artefact of the calibrator firing on a small number of dates where the baseline was holding a substantially negative-Sharpe position rather than a genuine OOS improvement.

\begin{table}[htb]
\centering
\small
\setlength{\tabcolsep}{3pt}
\resizebox{\columnwidth}{!}{%
\begin{tabular}{lrrr|rrr|r}
\toprule
& \multicolumn{3}{c|}{\textbf{Mid-cap}} & \multicolumn{3}{c|}{\textbf{Small-cap}} & \\
\cmidrule(lr){2-4}\cmidrule(lr){5-7}
\textbf{Model} & KMX & POOL & EXPE & CROX & SAM & BJRI & \textbf{mean} \\
\midrule
\llmPhi\,Phi-4-14B    & 0.27 & 0.15 & 0.08 & 0.00 & 0.14 & 0.17 & 0.14 \\
\llmQwen\,Qwen2.5-14B  & 0.45 & 0.90 & 1.18 & 0.38 & 0.80 & 1.00 & 0.79 \\
\llmLlama\,Llama-3.1-8B & 0.68 & 0.69 & 0.55 & 0.79 & 0.75 & 0.43 & 0.65 \\
\llmStarling\,Starling-7B  & 0.73 & 0.71 & 1.26 & 0.59 & 0.87 & 1.33 & 0.92 \\
\llmDeepSeek\,DeepSeek-7B  & 0.06 & 0.10 & 0.25 & 0.19 & 0.03 & 0.02 & 0.11 \\
\bottomrule
\end{tabular}
}
\caption{Per-ticker mean calibrated $\bar{\alpha}_{\text{IS}}$ across mid- and small-cap entities, all five backtested models. Compare with the mega-cap grid in Table~\ref{tab:exp2-alpha-grid}.}
\label{tab:exp2-midsmall-alpha-grid}
\end{table}

\paragraph{Mid- and small-cap robustness.}
Table~\ref{tab:honesty-drop} of the main paper extends each panel with one mid-cap (POOL) and one small-cap (SAM) probe ticker.
We additionally backtested every (model, ticker) combination across a six-ticker mid/small panel: KMX (CarMax), POOL (Pool Corporation), EXPE (Expedia) on the mid-cap side, and CROX (Crocs), SAM (Boston Beer), BJRI (BJ's Restaurants) on the small-cap side.
Tickers were chosen to have the full 2010--2025 daily price coverage that the IS+OOS protocol requires.
Table~\ref{tab:exp2-midsmall-alpha-grid} reports the per-(model, ticker) calibrated penalty $\bar{\alpha}_{\text{IS}}$.
Two patterns generalise from the mega-cap pool.
First, $\bar{\alpha}_{\text{IS}}$ scales with each model's overall memorisation strength (Phi-4 mid/small mean $0.14$; Qwen2.5-14B and Starling-7B both $\approx 0.9$; DeepSeek-7B-Chat $0.11$).
Second, the within-model spread on mid/small entities is non-trivial (Phi-4 max$-$min $0.27$; Qwen2.5-14B $0.80$), so the calibrator continues to discriminate among individual mid- and small-cap names rather than collapsing to a uniform setting.
The corresponding per-ticker IS+OOS backtests for all five models are in Table~\ref{tab:exp2-midsmall-detail}; for completeness this table includes the POOL and SAM rows already shown in the main paper for Phi-4-14B and Qwen2.5-14B alongside the remaining four mid/small tickers and the three other models.
Per-ticker IS deltas are noisier on small-caps because of lower trade volumes, smaller baseline portfolios, and the cascade of even small $\alpha$ firings through portfolio state; mean rows in the bottom of each panel capture the per-cap-class signal with less noise.

\begin{table}[!ht]
\centering
\small
\setlength{\tabcolsep}{3pt}
\resizebox{\columnwidth}{!}{%
\begin{tabular}{lrrrrrrrr}
\toprule
& \multicolumn{4}{c}{\textbf{IS (2010--2020)}} & \multicolumn{4}{c}{\textbf{OOS (2025)}} \\
\cmidrule(lr){2-5}\cmidrule(lr){6-9}
\textbf{Tkr} & B\$K & C\$K & $\Delta\%$ & $\Delta$Sh & B\$K & C\$K & $\Delta\%$ & $\Delta$Sh \\
\midrule
\multicolumn{9}{c}{\llmPhi\,\textit{Phi-4-14B (mid+small mean $\bar{\alpha}_{\text{IS}}{=}0.14$)}} \\
KMX  & 201 & 163 & $-$18.8 & $-$.07 &  44 &  47 & +5.5  & +.23 \\
POOL & 901 & 928 & +3.0    & +.00   &  67 &  74 & +9.8  & +.47 \\
EXPE & 165 & 206 & +24.8   & +.06   & 149 & 149 & 0.0   & .00 \\
CROX & 182 & 302 & +65.4   & +.12   &  59 &  73 & +23.4 & +.60 \\
SAM  & 692 & 562 & $-$18.8 & $-$.07 &  55 &  64 & +15.7 & +.84 \\
BJRI & 132 & 225 & +70.1   & +.16   & 114 & 109 & $-$4.4 & $-$.11 \\
\textit{mean} & & & \textbf{+21.0} & \textbf{+.03} & & & \textbf{+8.3} & \textbf{+.34} \\
\midrule
\multicolumn{9}{c}{\llmQwen\,\textit{Qwen2.5-14B (mid+small mean $\bar{\alpha}_{\text{IS}}{=}0.79$)}} \\
KMX  & 226 & 282 & +24.7   & +.08   &  55 &  64 & +17.2 & +.42 \\
POOL & 1027 & 753 & $-$26.7 & $-$.09 &  81 &  73 & $-$9.7  & $-$.33 \\
EXPE & 135 & 103 & $-$24.0 & $-$.11 & 140 & 136 & $-$3.4 & $-$.05 \\
CROX & 606 & 912 & +50.6   & +.10   &  58 &  98 & +69.7 & +1.11 \\
SAM  & 415 & 412 & $-$0.9  & +.01   &  69 &  76 & +10.2 & +.49 \\
BJRI & 117 & 132 & +12.8   & +.02   & 101 & 143 & +42.0 & +.96 \\
\textit{mean} & & & \textbf{+6.1} & \textbf{+.00} & & & \textbf{+21.0} & \textbf{+.43} \\
\midrule
\multicolumn{9}{c}{\llmLlama\,\textit{Llama-3.1-8B (mid+small mean $\bar{\alpha}_{\text{IS}}{=}0.65$)}} \\
KMX  & 421 & 106 & $-$74.9 & $-$.51 &  51 &  51 & 0.0    & .00 \\
POOL & 640 & 1017 & +58.8  & +.18   &  63 &  63 & 0.0    & .00 \\
EXPE & 198 & 184 & $-$7.1  & $-$.02 & 164 & 164 & 0.0    & .00 \\
CROX & 311 & 321 & +3.0    & +.01   &  63 &  79 & +25.9  & +.41 \\
SAM  & 502 & 444 & $-$11.6 & $-$.06 &  67 &  70 & +5.9   & +.29 \\
BJRI & 146 & 137 & $-$6.2  & $-$.02 & 135 & 106 & $-$21.4 & $-$.58 \\
\textit{mean} & & & \textbf{$-$6.3} & \textbf{$-$.07} & & & \textbf{+1.7} & \textbf{+.02} \\
\midrule
\multicolumn{9}{c}{\llmStarling\,\textit{Starling-7B (mid+small mean $\bar{\alpha}_{\text{IS}}{=}0.92$)}} \\
KMX  &  90 &  53 & $-$41.5 & $-$.42 &  61 &  72 & +18.5  & $-$.34 \\
POOL & 737 & 192 & $-$74.0 & $-$.58 &  77 & 100 & +30.7  & +1.30 \\
EXPE & 113 &  18 & $-$84.2 & $-$.74 & 117 &  82 & $-$30.4 & $-$1.38 \\
CROX & 189 & 209 & +10.8   & .00    &  89 &  95 & +6.4   & +.04 \\
SAM  & 335 & 129 & $-$61.6 & $-$.37 &  73 & 100 & +36.4  & +1.59 \\
BJRI & 152 &  75 & $-$50.7 & $-$.36 &  88 & 106 & +21.1  & +.62 \\
\textit{mean} & & & \textbf{$-$50.2} & \textbf{$-$.41} & & & \textbf{+13.8} & \textbf{+.31} \\
\midrule
\multicolumn{9}{c}{\llmDeepSeek\,\textit{DeepSeek-7B-Chat (mid+small mean $\bar{\alpha}_{\text{IS}}{=}0.11$)}} \\
KMX  &  97 & 120 & +24.1   & +.12   &  75 &  82 & +8.9   & +.84 \\
POOL & 289 & 314 & +8.5    & +.04   &  72 &  76 & +6.2   & +.32 \\
EXPE &  30 &  66 & +124.3  & +.43   & 106 & 121 & +14.6  & +.53 \\
CROX & 111 &  96 & $-$13.4 & $-$.03 &  72 &  72 & 0.0    & .00 \\
SAM  & 117 & 271 & +132.1  & +.41   &  71 &  75 & +5.3   & +.20 \\
BJRI & 101 &  26 & $-$74.6 & $-$.57 & 103 & 170 & +64.5  & +1.54 \\
\textit{mean} & & & \textbf{+33.5} & \textbf{+.07} & & & \textbf{+16.6} & \textbf{+.57} \\
\bottomrule
\end{tabular}
}
\caption{Full mid- and small-cap per-ticker results for all five backtested models. The POOL and SAM rows for Phi-4-14B and Qwen2.5-14B are reproduced here from Table~\ref{tab:honesty-drop} of the main paper for completeness. Daily rebalancing, $T{=}1.0$, seed 42. Per-ticker $\Delta$ is noisier on small-caps due to lower trade volumes and the cascade of small $\alpha$ firings through portfolio state; mean rows summarise each panel.}
\label{tab:exp2-midsmall-detail}
\end{table}

\subsection{Experiment 3: Cross-Model Roster}
\label{app:exp3-detail}

Table~\ref{tab:cross-model-models} lists the 11 instruction-tuned LLMs used in the cross-model alignment analysis (\S\ref{sec:exp-3}), with family, parameter count, and reported pre-training cutoff.
All reported cutoffs fall before January 2025, so the 2025 evaluation window follows every model's reported cutoff.

\begin{table}[!ht]
\centering
\small
\setlength{\tabcolsep}{3pt}
\begin{tabular}{llrl}
\toprule
\textbf{Model} & \textbf{Family} & \textbf{Size} & \textbf{Cutoff} \\
\midrule
\llmPhi\,Phi-4-14B            & Phi & 14B & Jun '24 \\
\llmLlama\,Llama-3.1-8B-Inst. & Llama & 8B & Mar '24 \\
\llmQwen\,Qwen2.5-14B-Inst.   & Qwen & 14B & Jun '24 \\
\llmQwen\,Qwen2.5-7B-Inst.    & Qwen & 7B & Jun '24 \\
\llmGLM\,GLM-4-9B-Chat        & GLM & 9B & Mar '24 \\
\llmMistral\,Mistral-7B-v0.3  & Mistral & 7B & May '24 \\
\llmYi\,Yi-1.5-9B-Chat        & Yi & 9B & May '24 \\
\llmOLMo\,OLMo-2-7B-Inst.     & OLMo & 7B & Nov '24 \\
\llmStarling\,Starling-LM-7B  & Starling & 7B & Mar '24 \\
\llmFalcon\,Falcon3-7B-Inst.  & Falcon & 7B & Sep '24 \\
\llmDeepSeek\,DeepSeek-7B-Chat & DeepSeek & 7B & Nov '23 \\
\bottomrule
\end{tabular}
\caption{Models for cross-model validation (11 models from 10 families, cutoffs before Jan 2025).}
\label{tab:cross-model-models}
\end{table}

\section{Per-Model Equity Curves}
\label{app:all-curves}

\begin{figure*}[htb]
\centering
\includegraphics[width=\textwidth]{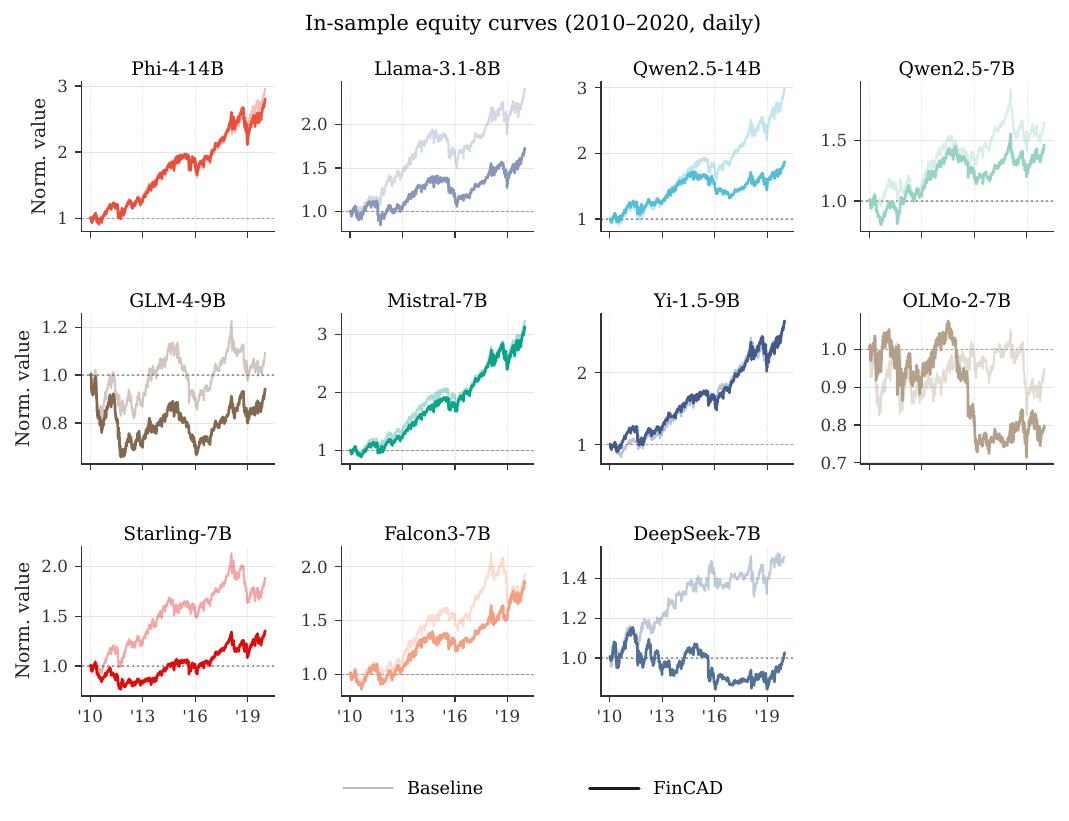}
\caption{In-sample SPY equity curves (2010--2020, daily) for all ten models. Faded: baseline decoding; bold: \ourmethod. Curves normalised to 1.0 at the start date.}
\label{fig:appendix-is-curves}
\end{figure*}

Figure~\ref{fig:alignment-scatter} in the main paper visualises the IS$\to$OOS alignment shift induced by \ourmethod via a per-model scatter of (IS, OOS) Sharpe pairs. Figure~\ref{fig:appendix-is-curves} complements this with the underlying in-sample equity curves for all ten models in the cross-model analysis. Within each panel, the faded line is baseline decoding and the bold line is \ourmethod. The gap between the two lines is a visual proxy for the look-ahead bias \ourmethod removes: it is near-zero for Phi-4, Llama-3.1, and Yi-1.5, but substantial for Qwen2.5-14B, Qwen2.5-7B, Starling-7B, and Falcon3-7B.

\section{Subset-Size Sensitivity}
\label{app:subset-sweep}

Table~\ref{tab:subset-sweep} reports the exhaustive subset analysis for Sharpe ratio across all subset sizes $k \in \{5, \ldots, 11\}$ from the $n{=}11$ models.
The observed difference $\Delta\rho = \bar{\rho}_{\text{C}} - \bar{\rho}_{\text{B}}$ remains between $+0.06$ and $+0.07$ across these subset sizes.
Because subsets at a given $k$ overlap, this analysis measures sensitivity to model-pool composition rather than sampling uncertainty over independent observations.

\begin{table}[!ht]
\centering
\small
\setlength{\tabcolsep}{4pt}
\begin{tabular}{rrrrr}
\toprule
$k$ & $\binom{11}{k}$ & $\bar{\rho}_{\text{B}}$ & $\bar{\rho}_{\text{C}}$ & $\Delta\rho$ \\
\midrule
5  & 462 & +.758 & +.818 & +.061 \\
6  & 462 & +.770 & +.834 & +.064 \\
7  & 330 & +.779 & +.846 & +.067 \\
8  & 165 & +.786 & +.855 & +.069 \\
9  &  55 & +.792 & +.862 & +.070 \\
10 &  11 & +.796 & +.868 & +.072 \\
11 &   1 & +.800 & +.873 & +.073 \\
\bottomrule
\end{tabular}
\caption{Descriptive Sharpe-ratio ranking alignment (Spearman $\rho$) across subset sizes. B = baseline; C = \ourmethod.}
\label{tab:subset-sweep}
\end{table}

\section{Artifact Licences, Provenance, and Attribution}
\label{app:licenses}

This section records the provenance, declared licence status, and experimental use of the principal datasets, codebases, and model weights. Dataset and model licences are reported from their official repository or model cards at the time of access.

\paragraph{Price data (FINSABER).}
The 2010--2025 daily price panel and the survivorship-, leakage-, and look-ahead-aware data conventions used throughout Experiments~2 and~3 are sourced from the FINSABER-V2 data release\footnote{\url{https://huggingface.co/datasets/finsaber-team/FINSABER-V2-Data}} \citep{li2025llmbasedfinancialinvestingstrategies}.
Its dataset card declares the licence as \textbf{other} and notes that users remain responsible for the terms of the underlying data providers. We use the data for academic evaluation and direct users to the upstream release; no price data are redistributed with \ourmethod.

\paragraph{General benchmark datasets.}
Experiment~1 uses \texttt{openai/gsm8k}, \texttt{TIGER-Lab/MMLU-Pro}, \texttt{HuggingFaceH4/MATH-500}, \texttt{openai/openai\_humaneval}, and FinEval through their Hugging Face releases.
The original \texttt{Salesforce/FinEval} dataset card\footnote{\url{https://huggingface.co/datasets/Salesforce/FinEval}} declares the \textbf{CC BY-NC 4.0 License} and notes that users remain responsible for the terms of its constituent source datasets. 
For reproducibility, we release the 20 multiple-choice subsets converted into the unified schema used by our evaluator. T
his derived release\footnote{\url{https://huggingface.co/datasets/waylonli/fineval-processed}} does not supersede the upstream licence or source-specific terms.
The dataset cards for GSM8K, MMLU-Pro, and HumanEval declare the \textbf{MIT License}. The MATH-500 card does not declare a licence and identifies OpenAI's MIT-licensed PRM800K repository as its immediate source. Except for the separate preprocessed FinEval release described above, these datasets are accessed for evaluation only and are not redistributed with \ourmethod.

\paragraph{Context-Aware Decoding.}
Our logit-blending update (Eq.~\ref{eq:cad}) follows the Context-Aware Decoding formulation of \citet{shi-etal-2024-trusting}, whose reference implementation is publicly available\footnote{\url{https://github.com/xhan77/context-aware-decoding}}.
We re-implement the decoding loop on top of HuggingFace Transformers in order to support (i) the entity-adaptive penalty (\S\ref{sec:adaptive-alpha}), (ii) the adversarial prior-discovery pipeline (\S\ref{sec:discovery}), and (iii) the adaptive-retry mechanism on JSON parse failure (Algorithm~\ref{alg:fincad}).
The reference repository did not contain a licence file when accessed. We use the published formulation as prior work and do not copy source files from the repository.

\paragraph{\ourmethod implementation.}
Our implementation is licensed under the \textbf{MIT License} and is released\footnote{\url{https://github.com/waylonli/FinCAD}}. The repository includes the inference and discovery code, experiment configurations, and discovery JSON artifacts; datasets and model weights remain available from their respective providers.

\paragraph{Single-stock backtesting harness (ai-hedge-fund).}
The structured-summary prompt template, JSON action schema, and execution-at-next-open conventions used in Experiments~2 and~3 are adapted from the open-source \texttt{ai-hedge-fund} project\footnote{\url{https://github.com/virattt/ai-hedge-fund}}, released under the \textbf{MIT License}.
We retain the agent-as-quantitative-analyst design but implement our own trading loop, a one-way transaction-cost model of 10~bps of trade notional (Appendix~\ref{app:backtesting-framework}), and the long-only single-stock constraint that defines the simulation envelope on top of HuggingFace Transformers and pandas.

\paragraph{Prompt optimisation (DSPy / MIPROv2).}
The adversarial prior-discovery pipeline (\S\ref{sec:discovery}) uses MIPROv2 \citep{khattab2024dspy} as released in DSPy under the \textbf{MIT License}.
We use DSPy as a library without modification and write a project-specific signature, metric, and proposer configuration as described in \S\ref{sec:discovery}.

\paragraph{Inference stack.}
Model inference uses HuggingFace Transformers (Apache License 2.0) and PyTorch (BSD-3-Clause License); ancillary data handling uses pandas (BSD-3-Clause).

\paragraph{Pre-trained model licenses.}
All eleven LLMs evaluated in \S\ref{sec:exp-3} are accessed through HuggingFace under their providers' stated terms; we use the pre-trained weights for non-commercial academic research consistent with each licence.
The majority (Qwen2.5-14B-Instruct, Qwen2.5-7B-Instruct, Mistral-7B-v0.3, Yi-1.5-9B-Chat, and OLMo-2-7B-Instruct) are released under the \textbf{Apache License 2.0}.
Phi-4-14B is released under the \textbf{MIT License}.
The remaining five models carry custom or community licences: Llama-3.1-8B-Instruct under the \textbf{Llama 3.1 Community Licence}, GLM-4-9B-Chat under the \textbf{GLM-4 Model Licence}, Falcon3-7B-Instruct under the \textbf{TII Falcon-LLM Licence 2.0}, DeepSeek-7B-Chat under the \textbf{DeepSeek Model Licence}, and Starling-LM-7B under \textbf{Apache 2.0 with a research-only restriction}.
Licences are as reported on each model's HuggingFace card at the time of writing.

\end{document}